%% file: main.tex
\def\isarxiv{1} 

\ifdefined\isarxiv
\documentclass[11pt]{article}

\usepackage[numbers]{natbib}

\else
\documentclass{article}
\usepackage{neurips_2022}
\fi

\usepackage{amsmath}
\usepackage{amsthm}
\usepackage{amssymb}
\usepackage{algorithm}
\usepackage{subfig}
\usepackage{algpseudocode}
\usepackage{graphicx}
\usepackage{grffile}
\usepackage{wrapfig,epsfig}
\usepackage{url}
\usepackage{xcolor}
\usepackage{epstopdf}
\usepackage{mathrsfs}

\usepackage{bbm}
\usepackage{dsfont}

\allowdisplaybreaks

\ifdefined\isarxiv

\usepackage{tikz}
\usepackage{hyperref}  
\hypersetup{colorlinks=true,citecolor=blue,linkcolor=blue} 
\usetikzlibrary{arrows}
\usepackage[margin=1in]{geometry}

\else

\usepackage{microtype}
\usepackage{hyperref}
\definecolor{mydarkblue}{rgb}{0,0.08,0.45}
\hypersetup{colorlinks=true, citecolor=mydarkblue,linkcolor=mydarkblue}

\fi

\newtheorem{theorem}{Theorem}[section]
\newtheorem{lemma}[theorem]{Lemma}
\newtheorem{definition}[theorem]{Definition}

\newtheorem{fact}[theorem]{Fact}

\newcommand{\wt}{\widetilde}
\newcommand{\ov}{\overline}

\newcommand{\R}{\mathbb{R}}

\renewcommand{\d}{\mathrm{d}}

\DeclareMathOperator*{\Z}{\mathbb{Z}}

\DeclareMathOperator{\poly}{poly}

\DeclareMathOperator{\nnz}{nnz}

\DeclareMathOperator{\rank}{rank}
\DeclareMathOperator{\diag}{diag}

\DeclareMathOperator{\reg}{reg}

\makeatletter
\newcommand*{\RN}[1]{\expandafter\@slowromancap\romannumeral #1@}
\makeatother

\usepackage{lineno}

\begin{document}

\ifdefined\isarxiv

\date{}

\title{Attention Scheme Inspired Softmax Regression}
\author{
Yichuan Deng\thanks{\texttt{ycdeng@cs.washington.edu}. The University of Washington.}
\and
Zhihang Li\thanks{\texttt{lizhihangdll@gmail.com}. Huazhong Agricultural University.}
\and
Zhao Song\thanks{\texttt{zsong@adobe.com}. Adobe Research.}
}

\else

\title{Intern Project} 
\maketitle 
\fi

\ifdefined\isarxiv
\begin{titlepage}
  \maketitle
  \begin{abstract}
\input{abstract}

  \end{abstract}
  \thispagestyle{empty}
\end{titlepage}

{
}
\newpage

\else

\begin{abstract}
\input{abstract}
\end{abstract}

\fi

\input{intro}

\input{tech}

\input{preli}

\input{loss}

\input{hessian}

\input{result}


\ifdefined\isarxiv
\bibliographystyle{alpha}
\bibliography{ref}
\else
\bibliography{ref}
\bibliographystyle{alpha}

\fi




\end{document}

%% file: abstract.tex
Large language models (LLMs) have made transformed changes for human society. One of the key computation in LLMs is the softmax unit. This operation is important in LLMs because it allows the model to generate a distribution over possible next words or phrases, given a sequence of input words. This distribution is then used to select the most likely next word or phrase, based on the probabilities assigned by the model. The softmax unit plays a crucial role in training LLMs, as it allows the model to learn from the data by adjusting the weights and biases of the neural network.  

In the area of convex optimization such as using central path method to solve linear programming. The softmax function has been used a crucial tool for controlling the progress and stability of potential function [Cohen, Lee and Song STOC 2019, Brand SODA 2020]. 

In this work, inspired the softmax unit, we define a softmax regression problem. Formally speaking, given a matrix $A \in \R^{n \times d}$ and a vector $b \in \R^n$, the goal is to use greedy type algorithm to solve 
\begin{align*}
    \min_{x} \| \langle \exp(Ax), {\bf 1}_n \rangle^{-1} \exp(Ax) - b \|_2^2.
\end{align*}
In certain sense, our provable convergence result provides theoretical support for why we can use greedy algorithm to train softmax function in practice.

%% file: intro.tex
\section{Introduction}

In the past few years, Large Language Models (LLMs) have experienced an explosive development. There is a series of results of LLMs, like Transformer \cite{vsp+17}, GPT-1 \cite{rns+18}, BERT \cite{dclt18}, GPT-2 \cite{rwc+19}, GPT-3 \cite{bmr+20}, PaLM \cite{cnd+22}, OPT \cite{zrg+22}. The success of a recent chatbot named ChatGPT \cite{cha22} by OpenAI has exemplified the use of LLMs in human-interaction tasks. Very recently, OpenAI released their new version of LLM, named GPT-4 \cite{o23}, which has been tested to perform much better even than previous ChatGPT \cite{bce+23}. These LLMs are  trained on massive amounts of textual data to generate natural language text. They have already shown their power on various real-work tasks, including natural language translation \cite{hwl21}, sentiment analysis \cite{uas20}, language modeling \cite{mms+19}, and even creative writing \cite{cha22, o23}. 

In the construction of the LLMs, \emph{attention} computation is a key component which is used to enhance the model's ability to focus on relevant parts of the input text \cite{vsp+17, rns+18, dclt18, rwc+19, bmr+20}. The attention matrix is defined as a squared matrix consisted of rows and columns related to the words or tokens, and the entries in the matrix represent the correlations between the words/tokens in the input text. The attention mechanism allows the model to selectively focus on specific parts of the input text when generating the output, rather than treating all input tokens equally. The attention mechanism is based on the idea that different parts of the input sequence contribute differently to the output sequence, and the model should learn to weigh these contributions accordingly. In LLMs, attention computation is typically implemented as a soft attention mechanism, where the weights are computed using a softmax function over the input sequence. The Attention computation can be described as follows (see \cite{zhdk23,as23,bsz23} as an example). 
\begin{definition}[Static Attention Computation]
    Given matrix $Q,K, V \in \R^{n \times d}$, we define
    \begin{align*}
        \mathsf{Att}(Q,K,V) := D^{-1} A V
    \end{align*}   
    where  $A := \exp(Q K^\top) \in \R^{n \times n}$ is a square matrix and  $D := \diag( A {\bf 1}_n ) \in \R^{n \times n}$ is a diagonal matrix.
\end{definition}
In the above definition, We use matrix $Q \in \R^{n \times d}$ to denote the query tokens, which are typically derived from the previous hidden state of the decoder. And we use matrix $K \in \R^{n \times d}$ and $V \in \R^{n \times d}$ to denote the key tokens and values. By the way we compute $A$, each entry of $A$ is computed as a dot product between the query vector $q$ and the key vector $k_i$, and the softmax function is applied to obtain the attention weights $A_{i, j}$. 

Motivated by the exp function in attention computation, previous work \cite{lsz23,gms23} has formally defined hyperbolic function (for example $f(x) = \exp(Ax), \cosh(Ax), \sinh(Ax)$) regression problems.
\begin{definition}[Hyperbolic Regression \cite{lsz23}]
Given $A \in \R^{n \times d}$ and $b \in \R^n$, the softmax regression problem is aiming for minimize the following objective function 
\begin{align*}
\min_{x \in \R^d} \| f(x) - b \|_2^2 .
\end{align*}
Here $f(x)$ can be either of $\exp(Ax)$, $\cosh(Ax)$ and $\sinh(Ax)$.
\end{definition}

In this work, we move one more step forward and to consider the normalization factor, $\langle f(x) , {\bf 1}_n \rangle^{-1}$ $= \langle \exp(Ax) , {\bf 1}_n \rangle^{-1}$. We will focus on the exp in the rest of the paper. Inspired by the softmax formulation in each row of the above attention computation, we formally define the softmax regression problem, 
\begin{definition}[Softmax Regression]
Given $A \in \R^{n \times d}$ and $b \in \R^n$, the softmax regression problem is aiming for minimize the following objective function 
\begin{align*}
\min_{x \in \R^d} \| \langle \exp(Ax) , {\bf 1}_n \rangle^{-1} \exp(Ax) - b \|_2^2 .
\end{align*}
\end{definition}
It is natural in practice to consider regularization \cite{llr23}, then we consider the regularized version of softmax regression. 
\begin{definition}[Regularized Softmax Regression]
Given $A \in \R^{n \times d}$, $b \in \R^n$, and $w \in \R^n$, the goal of the regularized softmax regression is to solve the following minimization problem,
\begin{align*}
\min_{x \in \R^d} 0.5 \cdot \| \langle \exp(Ax), {\bf 1}_n \rangle^{-1} \exp(Ax) - b \|_2^2 + 0.5 \cdot \| \diag(w) A x \|_2^2.
\end{align*}
\end{definition}

\subsection{Our Result}
Here we state our main result. We remark that since $\langle \exp(Ax), {\bf 1}_n \rangle^{-1} \exp(Ax)$ is always a probability distribution. Therefore, it is a natural to consider each entry in $b$ is nonnegative and its $\ell_1$ norm is at most $1$.
\begin{theorem}[Main Result, informal]
Given matrix $A \in \R^{n \times d}$, $b \in \R^n$, and $w \in \R^n$.

\begin{itemize}
    \item We define $f(x):=\langle \exp(Ax), \mathbf{1}_n \rangle^{-1} \exp(Ax)$.
    \item We use $x^*$ to denote the optimal solution of 
    \begin{align*}
    \min_{x \in \R^d} 0.5 \| f(Ax) - b \|_2^2 + 0.5 \| \diag(w) A x \|_2^2
    \end{align*}
    that
    \begin{itemize}
        \item $g(x^*) = {\bf 0}_d$.
        \item $\| x^* \|_2 \leq R$.
    \end{itemize}
    \item Suppose that $R \geq 10$.
    \item Assume that $\| A \| \leq R$. Here $\| A \|$ denotes the spectral norm of matrix $A$.
    
    \item Suppose that $b \geq {\bf 0}_{n}$ and $\| b \|_1 \leq 1$. Here ${\bf 0}_n$ denotes a length-$n$ vector where all the entries are zeros.

    \item Assume that $w_{i}^2 \geq 100 + l/\sigma_{\min}(A)^2$ for all $i \in [n]$. Here $\sigma_{\min}(A)$ denotes the smallest singular value of matrix $A$. 

    \item Let $M = n^{1.5} \exp(30 R^2)$.
    \item Let $l > 0$.
    \item Let $x_0$ denote an starting/initial point such that $M \| x_0 - x^* \|_2 \leq 0.1 l$.
    \item Let $\epsilon \in (0,0.1)$ be our accuracy parameter.
    \item Let $\delta \in (0,0.1)$ be our failure probability.
    \item Let $\omega$ denote the exponent of matrix multiplication.
\end{itemize}
There is a randomized algorithm (Algorithm~\ref{alg:main}) that
\begin{itemize}
\item  runs $\log(\| x_0 - x^* \|_2/ \epsilon)$ iterations 
\item spend  
\begin{align*}
O( (\nnz(A) + d^{\omega} ) \cdot \poly(\log(n/\delta)) 
\end{align*}
time per iteration,
\item and finally outputs a vector $\wt{x} \in \R^d$ such that
\begin{align*}
\Pr[ \| \wt{x} - x^* \|_2 \leq \epsilon ] \geq 1-\delta.
\end{align*}
\end{itemize}
\end{theorem}

We remark that, in previous work \cite{lsz23}, they only assume $\| b \|_2 \leq R$. The reason is in their setting, they don't consider the normalization parameter. It doesn't make sense for them to assume that $\| b \|_1 \leq 1$ because they're not trying to learn the distribution.

\paragraph{Roadmap.}

We organize the following paper as follows. 
In Section~\ref{sec:related}, we introduce some other projects that's related to or that has inspired our work. 
In Section~\ref{sec:overview} we provide a sketch for the techniques used in our project. 
In Section~\ref{sec:preli} we define the notations used in our work and provide some useful tools for exact algebra, approximate algebra and differential computation. 
In Section~\ref{sec:softmax} we provide detailed analysis of $L_{\exp}$, including its gradient and hessian. 
In Section~\ref{sec:psd} we proved that $L = L_{\exp} + L_{\reg}$ is a convex function. 
In Section~\ref{sec:lipschitz} we proved that the hessian of $L_{\exp}$ is Lipschitz. 
In Section~\ref{sec:newton} we provide an approximate version of newton method for solving convex optimization problem which is more efficient under certain assumptions. 
In Section~\ref{sec:result} we state our result of this paper and provide the algorithm for tackling the softmax regression problem.

\section{Related Work}\label{sec:related}

\subsection{Attention Theory}

\paragraph{Computation.}
Since the explosion of LLM, there have been a lot of theoretical works about the computation of attention \cite{kkl20, clp+21, zhdk23, as23, bsz23, lsz23, dms23}. 
Locality sensitive hashing (LSH) techniques have been employed in research to approximate attention. \cite{kkl20, clp+21, zhdk23}. Based on it, \cite{zhdk23} proposed KDEformer, an efficient approximation algorithm for the dot-product attention mechanism, with provable spectral norm bounds and superior performance on various pre-trained models. 
Recent research has investigated both static and dynamic approaches to attention computation \cite{as23, bsz23}. Additionally, \cite{lsz23} delved into regularized hyperbolic regression problems involving exponential, cosh, and sinh functions. \cite{dms23} proposed randomized and deterministic algorithms to sparsify the attention matrix in large language models, achieving high accuracy with significantly reduced feature dimension. 

\paragraph{Convergence and Optimization.}
There have been works trying to understanding attention computation on optimization and convergence perspective \cite{zkv+20, szks21, gms23, lsz23, llr23}. 
In practical attention models, adaptive methods often performs better than SGD. To understand this, \cite{zkv+20} showed that heavy-tailed distribution of the noise is one of the reason of the bad performance of SGD compared to adaptive methods, and provided new upper and lower bounds for convergence of adaptive methods under heavy-tailed noise in attention models. This answered the question of why adaptive methods performs better in attention models. 
\cite{szks21} explained why models sometimes attend to salient words and how the attention mechanism evolves throughout training, using a model property they defined, named Knowledge to Translate Individual Words (KTIW), which is learned early on from word co-occurrence statistics and later used to attend to input words while predicting the output. 
Recently, \cite{gms23} studied the regression problem inspired by the neural network with exponential activation function, and showed the convergence of a two-layer NN with large width (over-parameterized), while \cite{lsz23} focused on solving regularized exp, cosh and sinh regression problems inspired by Attention computation. \cite{llr23} explored how transformers learn the co-occurrence structure of words by examining attention-based network size, depth, and complexity through experiments and mathematical analysis, showing that the embedding and self-attention layers encode topical structure with higher average inner product and pairwise attention between same-topic words.

\paragraph{Privacy and Security.}
With the fast development of LLMs, the potential negative impact of abusing LLM has also been considered. To overcome this, without influencing the quality of the generated text, \cite{kgw+23} proposed a novel method to add watermark in LLM-generated text. The method needs no access to the parameters or API of the LLM. \cite{vkb23} introduced a formal definition of near access-freeness (NAF) and develops generative model learning algorithms to ensure that the model outputs do not resemble copyrighted data by more than $k$-bits, with experiments on language (transformers) and image (diffusion) generative models demonstrating strong protection against sampling protected content.

\subsection{Fast Linear Algebra}

\paragraph{Applications of Exponential Functions}

There are many theory problem use $\exp, \sinh, \cosh$ function as potential functions to prove the convergence of iterative optimization algorithms. In the work of \cite{cls19,b20,jswz21}, they use $\cosh$ function to define a potential function for measuring the central path. Such design can guarantee the central path method is robust and stable. Let $x \in \R^n$ denote the primal variables and let $s$ denote the slack variables of the central path algorithm. The central path is defined as tuple $(x,y,s,t)$ that satisfies
\begin{align*}
Ax & ~ = b, x > 0\\
A^\top y + s & ~ = c, s > 0\\
x_i s_i & ~ = t \mathrm{~for~all} i \in [n].
\end{align*}

Let $t$ denote the target at one step of central (also mathematically called the complementarity gap). The $xs$ can viewed as reality. In the ideal case, they hope $xs=t$. However, this is unlikely to happen. They are using the potential function $\Phi(xs) = \sum_{i=1}^n \cosh( x_is_i - t )$ to measure the difference between reality and target.

In \cite{qsz23}, they use $\cosh$ function to build a potential for rank-$1$ matrix sensing problem. Given a matrix $A \in \R^{d \times n}$, there are $n$ observations $x_i,y_i$ and $b_i = x_i^\top A y_i$. The goal of matrix sensing is to recover $A$ by using observations $\{ (x_i,y_i,b_i) \}_{i \in [n]}$.  They use the potential function $\Phi(x,y) = \sum_{i=1}^n \cosh( x_i^\top A y_i - b_i )$.

In standard linear $\ell_2$ regression, given matrix $A \in \R^{n \times d}$ and vector $b \in \R^n$, the formulation is usually $L(x) = \| A x - b \|_2^2 = (\sum_{i=1}^n (Ax)_i - b_i )^2$.  In \cite{lsz23}, they use $\cosh$ function to construct a $\ell_2$ loss such that $L(x) = \sum_{i=1}^n ( \cosh ( (Ax)_i ) - b_i )^2$. Furthermore, \cite{lsz23} also studied $\exp$ and $\sinh$ functions.

\paragraph{Sketching for Convex Optimization.}
Sketching technique has been widely-used in optimization problems such as  
linear programming \cite{cls19,jswz21,dly21,sy21,gs22}, empirical risk minimization \cite{lsz19,qszz23}, cutting plane method \cite{jlsw20}, computing John Ellipsoid \cite{ccly19,syyz22}, integral minimization problem \cite{jlsz23}, matrix completion \cite{gsyz23}, training over-parameterized neural tangent kernel regression \cite{bpsw21,szz21,z22,als+22}, matrix sensing \cite{dls23}.

%% file: tech.tex
\section{Technique Overview}\label{sec:overview}

Here in this section, we provide an overview of our techniques. 

\paragraph{Decomposition of Hessian for Softmax Regression.}
Recall the target function of our problem is in the form of
\begin{align*}
    \min_{x \in \R^d} 0.5\cdot\|f(Ax) - b\|_2^2 + 0.5 \cdot \|WAx\|_2^2,
\end{align*}
We divide the loss function with respect to above target function to the following two terms $L(x) := L_{\exp}(x) +  L_{\mathrm{reg}}(x)$, where
\begin{align*}
    L_{\exp}(x) & ~ := 0.5\cdot \|\langle\exp(Ax), \mathbf{1}_n\rangle^{-1}\cdot \exp(Ax) - b\|_2^2 \\
    L_{\mathrm{reg}}(x) & ~ := 0.5 \cdot \|WAx\|_2^2
\end{align*}
Calculating the Hessian of $L_{\exp}(x)$ directly is too complicated. To simplify this, we define two terms of $\alpha(x) := \langle\exp(Ax), \mathbf{1}_n\rangle$, $f(x) := \langle\exp(Ax), \mathbf{1}_n\rangle^{-1}\cdot \exp(Ax)$. Then to get the final Hessian to the loss functions, we calculate the Hessian step by step. To be specific, we divide the Hessian calculation into the following items: 
\begin{itemize}
    \item Hessian of $\exp(Ax)$;
    \item Hessian of $\alpha(x)$ and $\alpha^{-1}(x)$;
    \item Hessian of $f(x)$. 
\end{itemize}
After that, we notice a structured decomposition of Hessian of $L(x)$. We show that
\begin{align*}
        \frac{\d^2 L}{\d x_i^2} 
    & ~ = A_{*,i}^\top B(x)A_{*,i}\\
         \frac{\d^2 L}{\d x_i \d x_j} 
    & ~ = A_{*,i}^\top B(x) A_{*,j},
\end{align*}
 
where $B(x)$ is only function with $x$ and has no relation with respect to $i$ and $j$. In order to apply existing sparsification tool to boost the Hessian calculation (which is one of our main motivations), we construct specific decomposition to the two terms $B(x)$. We show that, $B$ can be viewed as sums of several rank-$1$ matrices and diagonal matrices. 

\paragraph{Hessian is Positive Definite.}
The key insight of this section lies in the analysis of volumetric barrier functions for solving semidefinite programming. \cite{a00,hjs+22}. 
With the decomposition of the Hessian matrix for $L_{\exp}$, the next step is to bound it. To be specific, by dividing $B(x)$ in the way of low-rank parts and diagonal parts, we can lower and upper bound each segment of them. And by combining them, we can get the bound for $B(x)$,
\begin{align*}
    -4I_n \preceq B(x) \preceq 8I_n.
\end{align*}
Now combine the Hessian for $L_{\exp}$ and $L_{\mathrm{reg}}(x)$ (Hessian for $L_{\mathrm{reg}}(x)$ is trivial $A^\top W^2A$) we get 
\begin{align*}
    \frac{\d^2 L}{\d x^2} = A^\top (B(x) + W^2)A.
\end{align*}
We show that, by assuming all $w_i^2$'s are lower bounded by $4 + l/\sigma_{\min}(A)^2$, the Hessian is positive definite $\frac{\d^2 L}{\d x^2} \succeq l \cdot I_d$. Further more, we show if all $w_i^2$'s are lower bounded by $100 + l/\sigma_{\min}(A)^2$, then the matrix $W^2$ can approximate the sum of $B(x) + W^2$ with a constant guarantee, i.e.,
\begin{align*}
    (1 - 1/10) \cdot (B(x) + W^2) \preceq W^2 \preceq (1 + 1/10) \cdot (B(x) + W^2).
\end{align*}
This allows us to apply sparsification tool on $W$ to approximate the Hessian. 

\paragraph{Lipschitz property for Hessian.}
The key insight of this section lies in the analysis of previous analysis for recurrent neural networks \cite{als19_dnn,als19_rnn}. 
By the above calculation of Hessian, we divide the Hessian matrix to different segments. Now with the decomposition (to be specific, we divide the Hessian into low-rank parts and diagonal parts), we show Lipschitz property for each term. We first show Lipschitz property for the basic terms: 
\begin{itemize}
    \item $\| \exp(Ax) \|_2 \leq \sqrt{n} \cdot \exp(R^2)$
    \item $\| \exp(Ax) - \exp(Ay) \|_2 \leq 2 \sqrt{n} \cdot R \exp(R^2) \cdot \| x - y \|_2$;
    \item $| \alpha(x) - \alpha(y) | \leq \sqrt{n} \cdot \| \exp(Ax) - \exp(Ay) \|_2$;
    \item $|\alpha(x)^{-1} - \alpha(y)^{-1}| \leq \beta^{-2} \cdot | \alpha(x) - \alpha(y) |$; (Later we will also prove an upper bound for $\beta^{-1}$, see Lemma~\ref{lem:beta})
    \item $\| f(x) - f(y) \|_2 \leq R_f \cdot \| x - y \|_2$. (Here $R_f$ is a function of $\beta^{-1}, \exp(R^2)$, see concrete definition in Lemma~\ref{lem:f_lipschitz})
\end{itemize}
Then, following the decomposition of the Hessian matrix, we show the Lipschitz property for each of the divided terms (we use $G_i$ for $i \in 1, \dots, 8$ to denote the terms) and combine them together to get the property of
\begin{align*}
    \| G_1 \| + \sum_{i=1}^8 \| G_i \| \leq 100 R \cdot \| f(x) - f(y) \|_2.
\end{align*}
With this property and a fact that $\|\frac{\d^2L}{\d x^2}(x) - \frac{\d^2L}{\d x^2}(y)\| \le \|A\| \cdot (\| G_1 \| + \sum_{i=1}^8 \| G_i \|) \cdot \|A\|$, by assuming any two points $x, y$ satisfy $\|x\|_2, \|y\|_2 \le R$ and $\|A(x-y)\|_\infty < 0.01$, we can show that the Hessian matrix is Lipschitz, i.e.,
\begin{align*}
    \|\frac{\d^2L}{\d x^2}(x) - \frac{\d^2L}{\d x^2}(y)\| \le \beta^{-2} n^{1.5}\exp(20R^2) \cdot \|x-y\|_2,
\end{align*}
for some small constant $\beta \in (0, 0.1)$, which implies the Lipschitz property for the Hessian. 

\paragraph{Approximated Newton Method with Sparsification Tool.}
Newton method is a widely-used and traditional tool used in optimization questions. In many optimization applications, computing $\nabla^2 L(x_t )$ or $(\nabla^2 L(x_t))^{-1}$ is quite expensive. Therefore, a natural motivation is to approximately formulate its Hessian or inverse of Hessian. In our setting, we want a faster implementation of Newton method. By above steps, we show our Hessian can be approximated by a matrix in the form of $A^\top D A$, where $D = W^2$ is a diagonal matrix. This inspires us to implement a standard tool \cite{dsw22, syyz22} that can generate a sparse matrix $\wt{D}$ such that 
\begin{align*}
    (1 - \epsilon) \cdot A^\top D A \preceq A^\top \wt{D} A \preceq (1 + \epsilon) \cdot A^\top D A
\end{align*}
in near input-sparsity time of $A$. By this tool, we can reduce the time for Hessian calculation of each iteration to the time of $\wt{O}(\nnz(A) + d^\omega)$. Here $\nnz(A)$ denotes the number of non-zero entries in matrix $A$. Let $\omega$ denote the exponent of matrix multiplication. Currently, $\omega \approx 2.373$ \cite{w12,lg14,aw21}.

%% file: preli.tex
\section{Preliminary}\label{sec:preli}

In this section, we provide the preliminaries used in our paper. In Section~\ref{sec:preli:notations} we introduce the notations we use. In Section~\ref{sec:preli:basic_algebras} we provide some basic facts for exact computation. In Section~\ref{sec:preli:vec_norm} we provide some tools for finding the bound of norms based on vectors. In Section~\ref{sec:preli:mat_norm} we provide some tools for finding the bound of norms related to matrices. In Section~\ref{sec:preli:psd}, we provide basic inequalities for psd matrices. In Section~\ref{sec:preli:basic_derivative}, we state several basic rules for calculus. In Section~\ref{sec:preli:reg} we provide the regularization term $L_{\reg}$ and compute $\nabla L_{\reg}$ and $\nabla^2 L_{\reg}$.

\subsection{Notations}\label{sec:preli:notations}
We denote the $\ell_p$ norm of a vector $x$ by $\| x \|_p$, i.e., $\| x \|_1 := \sum_{i=1}^n |x_i|$, $\| x \|_2 := (\sum_{i=1}^n x_i^2)^{1/2}$ and $\| x \|_{\infty}:= \max_{i \in [n]} |x_i|$. 
For a vector $x \in \R^n$, $\exp(x) \in \R^n$ denotes a vector where $\exp(x)_i$ is $\exp(x_i)$ for all $i \in [n]$.
For $n > k$, for any matrix $A \in \R^{n \times k}$, we denote the spectral norm of $A$ by $\| A \|$, i.e., $\| A \| := \sup_{x\in\R^k} \| A x \|_2 / \| x \|_2$. We use $\sigma_{\min}(A)$ to denote the minimum singular value of $A$. 
Given two vectors $x, y \in \R^n$, we use $\langle x, y\rangle$ to denote $\sum_{i=1}^n x_i y_i$. 
Given two vectors $x,y \in \R^n$, we use $x \circ y$ to denote a vector that its $i$-th entry is $x_iy_i$ for all $ i \in [n]$.
We use $e_i \in \R^n$ to denote a vector where $i$-th entry is $1$ and all the other entries are $0$. 
Let $x \in \mathbb{R}^n$ be a vector. We define $\mathrm{diag}(x) \in \mathbb{R}^{n \times n}$ as the diagonal matrix whose diagonal entries are given by $\mathrm{diag}(x)_{i,i} = x_i$ for $i = 1,\ldots,n$, and all off-diagonal entries are zero. 
For a symmetric matrix $A \in \R^{n \times n}$, we say $A \succ 0$ (positive definite (PD) ), if for all $x \in \R^n \backslash \{ {\bf 0}_n \}$, we have $x^\top A x > 0$. 
For a symmetric matrix $A \in \R^{n \times n}$, we say $A \succeq 0$ (positive semidefinite (PSD) ), if for all $x \in \R^n$,  we have $x^\top A x \geq 0$. The Taylor Series for $\exp(x)$ is $\exp(x) = \sum_{i=0}^{\infty}\frac{x^i}{i!}$. 

\subsection{Basic Algebras}\label{sec:preli:basic_algebras}

\begin{fact}\label{fac:basic}
For vectors $u, v, w \in \R^n$. We have
\begin{itemize}
    \item $\langle u, v \rangle = \langle u \circ v, {\bf 1}_n \rangle$
    \item $\langle u \circ v, w \rangle =  \langle u \circ v \circ w, {\bf 1}_n \rangle$
    \item $\langle u,v \rangle = \langle v,u \rangle$
    \item $\langle u,v\rangle  = u^\top v = v^\top u$
\end{itemize}
\end{fact}

\begin{fact}\label{fac:circ_diag}
For any vectors $u, v, w \in \R^n$, we have 
\begin{itemize}
    \item $u \circ v = v \circ u = \diag(u) \cdot v = \diag(v) \cdot u $
    \item $u^\top (v \circ w)= u^\top \diag(v) w$
    \item $u^\top (v \circ w) = v^\top (u \circ w) = w^\top (u \circ v)$
    \item $u^\top \diag(v) w = v^\top \diag(u) w = u^\top \diag(w) v$
    \item $\diag(u) \cdot \diag(v) \cdot {\bf 1}_n = \diag(u) v$
    \item $\diag(u \circ v) = \diag(u) \diag(v)$
    \item $\diag(u) + \diag(v) = \diag(u + v) $
\end{itemize}
\end{fact}

\subsection{Basic Vector Norm Bounds}\label{sec:preli:vec_norm}

\begin{fact}\label{fac:vector_norm}
For vectors $u, v \in \R^n$, we have
\begin{itemize}
    \item $\langle u, v \rangle \leq \| u \|_2 \cdot \| v \|_2$ (Cauchy-Schwarz inequality)
    \item $\|\diag(u)\| \leq \|u\|_{\infty}$
    \item $\| u \circ v \|_2 \leq \| u \|_{\infty} \cdot \| v \|_2$
    \item $\| u \|_{\infty} \leq \| u \|_2 \leq \sqrt{n} \cdot \|u\|_\infty$
    \item $\|u\|_2 \leq \|u\|_1 \leq \sqrt{n}\cdot \| u \|_2$
    \item $\| \exp(u) \|_{\infty} \leq  \exp( \| u \|_{\infty}) \leq \exp(\| u \|_2)$ 
    \item Let $\alpha$ be a scalar, then $\| \alpha \cdot u \|_2 = |\alpha| \cdot \| u \|_2$
    \item $\|u + v\|_2 \leq \|u\|_2 + \|v\|_2$.
    \item For any $\| u - v \|_{\infty} \leq 0.01$, we have $\| \exp(u) - \exp(v) \|_2 \leq \| \exp(u) \|_2 \cdot 2 \| u - v \|_{\infty}$
\end{itemize}
\end{fact}
\begin{proof}
    For all the other facts we omit the details. We will only prove the last fact.
    
    We have
    \begin{align*}
    \| \exp(u) - \exp(v) \|_2 = & ~ \| \exp(u) \circ ( {\bf 1}_n - \exp(v-u) ) \|_2 \\
    \leq & ~ \| \exp(u) \|_2 \cdot \| {\bf 1}_n - \exp(v - u) \|_{\infty} \\
    \leq & ~ \| \exp(u) \|_2 \cdot 2 \| u - v \|_{\infty},
    \end{align*}
    where the 1st step follows from definition of $\circ$ operation and $\exp()$, the 2nd step follows from $\| u \circ v \|_2 \leq \| u \|_{\infty} \cdot \| v \|_2$, the 3rd step follows from $| \exp(x) -1 | \leq 2x$ for all $x \in (0,0.1)$.

\end{proof}

\subsection{Basic Matrix Norm Bounds}\label{sec:preli:mat_norm}

\begin{fact}\label{fac:matrix_norm}
For matrices $U,V$, we have 
\begin{itemize}
    \item $\| U^\top \| = \| U \|$
    \item $\| U \| \geq \| V \| - \| U - V \|$
    \item $\| U + V \| \leq \| U \| + \| V \|$
    \item $\| U \cdot V \| \leq \| U \| \cdot \| V \|$ 
    \item If $U \preceq \alpha \cdot V$, then $\| U \| \leq \alpha \cdot \| V \|$
    \item For scalar $\alpha \in \R$, we have $\| \alpha \cdot U \| \leq |\alpha| \cdot \| U \|$
    \item For any vector $v$, we have $\| U v \|_2 \leq \| U \| \cdot \| v \|_2$.
    \item Let $u, v \in \R^n$ denote two vectors, then we have $\| u v^\top \| \leq \| u \|_2 \| v \|_2$
\end{itemize}
\end{fact}

\subsection{Basic PSD}\label{sec:preli:psd}

\begin{fact}\label{fac:psd}
    Let $u, v \in \R^n$, We have:
    \begin{itemize}
        \item $u u^\top \preceq \|u\|_2^2 \cdot I_n$.
        \item $\diag(u) \preceq \|u\|_2 \cdot I_n$
        \item $\diag(u \circ u) \preceq \|u\|_2^2 \cdot I_n$
        \item $uv^\top + v u^\top \preceq u u^\top + v v^\top$
        \item $uv^\top + v u^\top \succeq -( u u^\top + v v^\top )$
        \item $(v \circ u) (v \circ u )^\top \preceq \| v \|_{\infty}^2 u u^\top$
    \end{itemize}
\end{fact}

\subsection{Basic Derivative Rules}\label{sec:preli:basic_derivative}

\begin{fact}\label{fac:basic_derivative}
Let $f$ be a differentiable function.

We have
\begin{itemize}
    \item Part 1. $\frac{\d }{\d x} \exp(x) = \exp(x)$
    \item Part 2. For any $j \neq i$, $\frac{\d }{\d x_i} f(x_j) = 0$
\end{itemize}
\end{fact}

\begin{fact}[Rules of differentiation]\label{fac:derivative_rules}
Let $f$ denote a differentiable function. 

For all $n, i \in \Z_+$, we have
\begin{itemize}
    \item Sum rule 1. $\frac{\d}{\d t} \sum_{l = 1}^n f(x_l) = \sum_{l=1}^n \frac{\d}{\d t} f(x_i)$
    \item Sum rule 2. $\frac{\d}{\d x_i} \sum_{l = 1}^n f(x_l) = \frac{\d}{\d x_i} f(x_i)$
    \item Chain rule. $\frac{\d}{\d x_i} f(g(x_i)) = f'(g(x_i)) \cdot g'(x_i)$
    \item Difference rule. $\frac{\d}{\d x_i} (f(x_i) - g(x_i)) = \frac{\d}{\d x_i} f(x_i) - \frac{\d}{\d x_i} g(x_i)$
    \item Product rule. $\frac{\d}{\d x_i}(f(x_i)g(x_i)) = f'(x_i)g(x_i) + f(x_i)g'(x_i)$
    \item Constant multiple rule. For any $x \neq y$, $\frac{\d }{\d x_i} (y_i \cdot f(x_i)) = y_i \cdot \frac{\d }{\d x_i} f(x_i)$
\end{itemize}

\end{fact}

\subsection{Regularization}\label{sec:preli:reg}

\begin{definition}\label{def:L_reg}
Given matrix $A \in \R^{n \times d}$.
For a given vector $w \in \R^n$, let $W = \diag(w)$. 
We define $L_{\reg} : \R^d \rightarrow \R$ as follows
\begin{align*}
L_{\reg}(x):= 0.5 \| W A x\|_2^2
\end{align*}
\end{definition}

\begin{lemma}[Folklore, see \cite{lsz23} as an example]\label{lem:L_reg_gradient_hessian}
For a given vector $w \in \R^n$, let $W = \diag(w)$. Let $L_{\reg} : \R^d \rightarrow \R$ be defined as Definition~\ref{def:L_reg}.

Then, we have
\begin{itemize}
\item The gradient is
\begin{align*}
\frac{\d L_{\reg}}{ \d x} = A^\top W^2 Ax
\end{align*}
\item The Hessian is
\begin{align*}
\frac{\d^2 L_{\reg}}{ \d x^2} = A^\top W^2 A
\end{align*}
\end{itemize}
\end{lemma}

%% file: loss.tex
\section{Softmax Regression Loss}\label{sec:softmax}

In this section, we provide detailed computation for $\nabla L_{\exp}$ and $\nabla^2 L_{\exp}$. 
In Section~\ref{sec:softmax:def}, we define $f(x)$ and $\alpha(x)$ to simplify the computation for $\nabla L_{\exp}$ and $\nabla^2 L_{\exp}$. 
In Section~\ref{sec:softmax:gradient}, we compute $\nabla L_{\exp}$ step by step.  
In Section~\ref{sec:softmax:loss_gradient_lipschitz}, we define the gradient of Loss function and also prove the Lipschitz property for gradient. 
In Section~\ref{sec:softmax:hessian_step_1}-\ref{sec:softmax:hessian_step_5}, we compute $\nabla^2 L_{\exp}$ step by step. 
To be specific, 
in Section~\ref{sec:softmax:hessian_step_1}, we compute $\nabla^2 \exp(Ax)$; 
in Section~\ref{sec:softmax:hessian_step_2}, we compute $\nabla^2 \alpha(x)$; 
in Section~\ref{sec:softmax:hessian_step_3}, we compute $\nabla^2 \alpha(x)^{-1}$; 
in Section~\ref{sec:softmax:hessian_step_4}, we compute $\nabla^2 f(x)$; 
in Section~\ref{sec:softmax:hessian_step_5}, we compute $\nabla^2 L_{\exp}$.
In Section~\ref{sec:softmax:help}, we provide some result to aid the computation in Section~\ref{sec:softmax:decompose}.
In Section~\ref{sec:softmax:decompose}, we split $\nabla^2 L_{\exp}$ into several low rank matrices and diagonal matrices.

\subsection{Definitions}\label{sec:softmax:def}

We define function softmax $f$ as follows
\begin{definition}[Function $f$]\label{def:f}
Given a matrix $A \in \R^{n \times d}$. Let ${\bf 1}_n$ denote a length-$n$ vector that all entries are ones.  
We define prediction function $f: \R^d \rightarrow \R^n$ as follows 
\begin{align*}
f(x) := \langle \exp(Ax) , {\bf 1}_n \rangle^{-1} \cdot \exp(Ax) .
\end{align*}
\end{definition}
Then we have
\begin{lemma}\label{lem:basic_f}
Let $f : \R^d \rightarrow \R^n$ 
be defined as Definition~\ref{def:f}, then we have for all $x \in \R^d$,  
\begin{itemize}
    \item $\| f(x) \|_2 \leq \| f(x) \|_1 \leq 1$.
    \item $0 \preceq f(x) f(x)^\top \preceq I_n$.
    \item For any vector $b$, $0 \preceq (b \circ f(x)) (b\circ f(x))^\top \preceq \| b \|_{\infty}^2 f(x) f(x)^\top \preceq \| b \|_{\infty}^2 I_n$
    \item For any vector $b$, $\diag (b \circ b) \preceq \| b \|_{\infty}^2 I_n$
    \item $0 \preceq \diag(f(x)) \preceq \| f(x) \|_{\infty} I_n \preceq  \| f(x) \|_2 I_n$.
    \item $0 \preceq \diag(f(x) \circ f(x)) \preceq \| f(x) \|_{\infty}^2 I_n \preceq \| f(x) \|_2 I_n$.
\end{itemize}
\end{lemma}
\begin{proof}
The proofs are very straightforward, so we omitted the details here.
\end{proof}

\begin{definition}[Loss function $L_{\exp}$]\label{def:L_exp}
Given a matrix $A \in \R^{n \times d}$ and a vector $b \in \R^n$. 
We define loss function $L_{\exp} : \R^d \rightarrow \R$ as follows 
\begin{align*}
L_{\exp}(x) := 0.5 \cdot \| \langle \exp(Ax) , {\bf 1}_n \rangle^{-1} \exp(A x) - b \|_2^2.
\end{align*}
\end{definition}

For convenient, we define two helpful notations $\alpha$ and $c$
\begin{definition}[Normalized coefficients]
\label{def:alpha}
    We define $\alpha : \R^d \rightarrow \R$ as follows
    \begin{align*}
    \alpha(x) := \langle \exp(Ax), {\bf 1}_n \rangle.
    \end{align*}
    Then, we can rewrite $f(x)$ (see Definition~\ref{def:f}) and $L_{\exp}(x)$ (see Definition~\ref{def:L_exp}) as follows
    \begin{itemize}
        \item $f(x) = \alpha(x)^{-1} \cdot \exp(Ax)$.
        \item $L_{\exp}(x) = 0.5 \cdot \| \alpha(x)^{-1} \cdot \exp(Ax) - b \|_2^2$.
        \item $L_{\exp}(x) = 0.5 \cdot \| f(x) - b \|_2^2$.
    \end{itemize}
\end{definition}

\begin{definition}
\label{def:c}
    We define function $c: \R^d \in \R^n$ as follows
    \begin{align*}
    c(x) := f(x) - b.
    \end{align*}
    Then we can rewrite $L_{\exp}(x)$ (see Definition~\ref{def:L_exp}) as follows
    \begin{itemize}
        \item $L_{\exp}(x) = 0.5 \cdot \| c(x) \|_2^2$.
    \end{itemize}
\end{definition}

\subsection{Gradient}\label{sec:softmax:gradient}

\begin{lemma}[Gradient]\label{lem:gradient_computation}
If the following conditions hold
\begin{itemize}
    \item Given matrix $A \in \R^{n \times d}$ and a vector $b \in \R^n$.
    \item Let $\alpha(x)$ be defined in Definition~\ref{def:alpha}.
    \item Let $f(x)$ be defined in Definition~\ref{def:f}.
    \item Let $c(x)$ be defined in Definition~\ref{def:c}.
    \item Let $L_{\exp}(x)$ be defined in Definition~\ref{def:L_exp}.
\end{itemize}
 For each $i \in [d]$, we have
\begin{itemize}
    \item Part 1. 
    \begin{align*}
        \frac{ \d \exp(Ax)}{ \d x_i} = \exp(Ax) \circ A_{*,i}
    \end{align*}
    \item Part 2.
    \begin{align*}
        \frac{\d \langle \exp(Ax) , {\bf 1}_n \rangle }{\d x_i} = \langle \exp(Ax) , A_{*,i}\rangle
    \end{align*}
    \item Part 3.
    \begin{align*}
        \frac{\d \alpha(x)^{-1} }{\d x_i} = -\alpha(x)^{-1} \cdot \langle f(x), A_{*,i} \rangle
    \end{align*}
    \item Part 4.
    \begin{align*}
        \frac{\d f(x) }{\d x_i} 
        = \frac{ \d c(x) }{ \d x_i } =  & ~ -  \langle f(x), A_{*,i}\rangle \cdot f(x) + f(x) \circ A_{*,i}
    \end{align*}   
    \item Part 5. 
    \begin{align*}
        \frac{\d \langle f(x), A_{*,i} \rangle}{ \d x_i} = - \langle f(x), A_{*,i} \rangle^2 + \langle f(x), A_{*,i} \circ A_{*,i} \rangle
    \end{align*}
    \item Part 6. For each $j \neq i$
    \begin{align*}
        \frac{\d \langle f(x), A_{*,i} \rangle}{ \d x_j} = - \langle f(x), A_{*,i} \rangle \cdot \langle f(x), A_{*,j} \rangle  + \langle f(x), A_{*,i} \circ A_{*,j} \rangle
    \end{align*}
    \item Part 7.  
    \begin{align*}
        \frac{\d L_{\exp} (x) }{\d x_i} =  A_{*,i}^\top \cdot (f(x)(f(x) - b)^\top f(x) + \diag(f(x))(f(x) - b))
    \end{align*}
\end{itemize}
\end{lemma}
\begin{proof}
    {\bf Proof of Part 1.}
    For each $i \in [d]$, we have  
    \begin{align*}
     \frac{ \d ( \exp(Ax) )_i }{ \d x_i } 
     = & ~ \exp(Ax)_i \cdot \frac{\d (Ax)_i}{\d x_i} \\
     = & ~ \exp(Ax)_i \cdot \frac{ (A \d x)_i}{\d x_i} \\
     = & ~ \exp(Ax)_i \cdot {A_{*,i}}
    \end{align*}
    where the 1st step follows from simple algebra, the 2nd step follows from Fact~\ref{fac:basic_derivative}, the 3rd step follows from simple algebra.

    Thus, we have
    \begin{align*}
        \frac{ \d \exp(Ax)}{ \d x_i} = \exp(Ax) \circ A_{*,i}
    \end{align*}
    
    {\bf Proof of Part 2.}
    It trivially follows from arguments in Part 1.
    
    {\bf Proof of Part 3.}
    \begin{align*}
        \frac{\d \alpha(x)^{-1} }{ \d x_i}
        = & ~ \frac{\d \langle \exp(Ax) , {\bf 1}_n \rangle^{-1} }{\d x_i } \\
        = & ~ -1 \cdot \langle \exp(Ax) , {\bf 1}_n \rangle^{-1 - 1} \cdot \frac{\d}{\d x_i}(\langle \exp(Ax) , {\bf 1}_n \rangle) \\
        = & ~ -\langle \exp(Ax) , {\bf 1}_n \rangle^{-2} \langle \exp(Ax) , A_{*,i} \rangle \\
        = & ~ - \alpha(x)^{-1} \langle f(x), A_{*,i} \rangle
    \end{align*}
    where the 1st step follows from $\frac{\d y^{z}}{\d x}=z \cdot y^{z-1} \frac{\d y}{\d x}$, the 2nd step follows from results in Part 2.

    {\bf Proof of Part 4.}
    \begin{align*}
        \frac{\d f(x)}{ \d x_i}
        = & ~ \frac{\d \langle \exp(Ax) , {\bf 1}_n \rangle^{-1} \exp(Ax) }{\d x_i} \\
        = & ~ \exp(Ax) \cdot \frac{\d}{\d x_i}(\langle \exp(Ax) , {\bf 1}_n \rangle^{-1}) + \langle \exp(Ax) , {\bf 1}_n \rangle^{-1} \cdot \frac{\d}{\d x_i}\exp(Ax) \\
        = & ~ - \langle \exp(Ax) , {\bf 1}_n \rangle^{-2} \cdot \langle \exp(Ax) , A_{*,i}  \rangle \cdot \exp(Ax) \\
        & ~ +  \langle \exp(Ax) , {\bf 1}_n \rangle^{-1} \cdot \exp(Ax) \circ A_{*,i} \\
        = & ~ - \langle f(x), A_{*,i} \rangle \cdot f(x) + f(x) \circ A_{*,i}
    \end{align*}
    where the 1st step follows from Definition of $f$, 2nd step follows from differential chain rule, the 3rd step follows from the result from Part 2 and Part 3, the forth step follows from definition of $f$ (see Definition~\ref{def:f}).

    {\bf Proof of Part 5}

    \begin{align*}
        \frac{\d \langle f(x), A_{*,i} \rangle}{\d x_i}
        = & ~ A_{*,i}^\top\frac{\d f(x)}{\d x_i} \\
        = & ~ A_{*,i}^\top (- \langle f(x), A_{*,i} \rangle \cdot f(x) + f(x) \circ A_{*,i}) \\
        = & ~ -\langle f(x), A_{*,i} \rangle \cdot A_{*,i}^\top f(x) + A_{*,i}^\top f(x) \circ A_{*,i} \\
        = & ~ - \langle f(x), A_{*,i} \rangle^2 + \langle f(x), A_{*,i} \circ A_{*,i} \rangle
    \end{align*}
    where the 1st step follows from extracting $A_{*,i}$, the 2nd step follows from result of Part 4, the 3rd step follows from simple algebra, the last step follows from simple algebra.

    {\bf Proof of Part 6.}

    \begin{align*}
        \frac{\d \langle f(x), A_{*,i} \rangle}{\d x_j}
        = & ~ A_{*,i}^\top\frac{\d f(x)}{\d x_j} \\
        = & ~ A_{*,i}^\top (- \langle f(x), A_{*,j} \rangle \cdot f(x) + f(x) \circ A_{*,j}) \\
        = & ~ -\langle f(x), A_{*,j} \rangle \cdot A_{*,i}^\top f(x) + A_{*,i}^\top f(x) \circ A_{*,j} \\
        = & ~ -\langle f(x), A_{*,j} \rangle\langle f(x), A_{*,i} \rangle + \langle A_{*,i}, f(x) \circ A_{*,j} \rangle \\
        = & ~ - \langle f(x), A_{*,i} \rangle \cdot \langle f(x), A_{*,j} \rangle  + \langle f(x), A_{*,i} \circ A_{*,j} \rangle
    \end{align*}
    where the 1st step follows from extracting $A_{*,i}$, the 2nd step follows from result of Part 4, the 3rd step follows from simple algebra, the 4th step follows from simple algebra, the last step follows from simple algebra.

    {\bf Proof of Part 7.}
    \begin{align*}
        \frac{\d L_{\exp}(x)}{\d x_i}
        = & ~ \frac{\d}{\d x_i}(0.5 \cdot \| f(x) - b \|_2^2) \\
        = & ~ ( f(x)- b)^\top \frac{\d}{\d x_i} ( f(x) - b) \\
        = & ~ ( f(x) - b)^\top ( - \langle f(x), A_{*,i} \rangle \cdot f(x) + f(x) \circ A_{*,i}) \\
        = & ~ A_{*,i}^\top f(x)(f(x) - b)^\top f(x) + (f(x) - b)^\top f(x) \circ A_{*,i} \\
        = & ~ A_{*,i}^\top f(x)(f(x) - b)^\top f(x) + A_{*,i}^\top f(x) \circ (f(x) - b) \\
        = & ~ A_{*,i}^\top (f(x)(f(x) - b)^\top f(x) + \diag(f(x))(f(x) - b))
    \end{align*}
    where the 1st step follows from the definition of $f$, the 2nd step follows from $\frac{\d \|y\|^2_2}{\d x} = 2y^\top \frac{\d y}{\d x}$, the 3rd step follows from the result in Part 4, the forth step follows from simple algebra, the 5th step step  follows from simple algebra, and the last step follows from simple algebra.

\end{proof}
 
\subsection{Definition of Gradient}\label{sec:softmax:loss_gradient_lipschitz}
In this section, we use $g(x)$ to denote the gradient of $L_{\exp}(x)$.
\begin{definition}\label{def:g}
If the following conditions hold
\begin{itemize}
\item Let $L_{\exp}(x)$ be defined as Definition~\ref{def:L_exp}.
\item Let $c(x)$ be defined as Definition~\ref{def:c}.
\item Let $f(x)$ be defined as Definition~\ref{def:f}.
\end{itemize} 
We define $g(x) \in \R^d$ as follows
\begin{align*}
g(x):= \underbrace{A^\top }_{d \times n} \cdot \Big( \underbrace{f(x) }_{n \times 1} \underbrace{ \langle c(x), f(x) \rangle }_{ \mathrm{scalar} } + \underbrace{ \diag(f(x)) }_{ n\times n} \underbrace{ c(x) }_{n \times 1} \Big) 
\end{align*}
Equivalently, for each $i \in [d]$, we define 
\begin{align*}
g(x)_i := \underbrace{ \langle A_{*,i}, f(x) \rangle }_{\mathrm{scalar}} \cdot \underbrace{ \langle c(x), f(x) \rangle }_{ \mathrm{scalar} } + \underbrace{ \langle A_{*,i}, f(x) \circ c(x)\rangle }_{ \mathrm{scalar} }.  
\end{align*}
\end{definition}

\begin{lemma}\label{lem:gradient_lipschitz}
If the following conditions hold
\begin{itemize}
    \item Let $g_1 : \R^d \rightarrow \R^n$ be defined as $g_1(x):= f(x) \langle c(x), f(x) \rangle$
    \item Let $g_2 : \R^d \rightarrow \R^n$  be defined as $g_2(x): = \diag( f(x) ) c(x)$
    \item Let $R_f$ be parameter such that 
    \begin{itemize}
        \item $\| f(x) - f(y) \|_2 \leq R_f \cdot \| x - y \|_2$
        \item $\| c(x) - c(y) \|_2 \leq R_f \cdot \| x - y \|_2$
    \end{itemize}
    \item Let $R_{\infty} \in (0,2]$ be parameter such that
    \begin{align*}
        R_{\infty}:=\max\{ \| f(x) \|_2, \| f(y) \|_2, \| c(x) \|_2, \| c (y ) \|_2 \}
    \end{align*}
\end{itemize}
We can show
\begin{itemize}
\item Part 1.
\begin{align*}
\| g_1(x) - g_1(y) \|_2 \leq 3 R_f R_{\infty}^2  \| x - y \|_2 
\end{align*}
\item Part 2.
\begin{align*}
    \| g_2(x) - g_2(x) \|_2 \leq 2 R_f R_{\infty} \| x- y \|_2
\end{align*}
\item Part 3.
\begin{align*}
    \| g_1(x) + g_2 (x) - g_1(y) - g_2(y) \|_2 \leq 8 R_f R_{\infty} \| x- y \|_2
\end{align*}
\item Part 4.
\begin{align*}
    \| g(x) - g(y) \|_2 \leq 8 \cdot \| A \| \cdot R_f \cdot R_{\infty} \| x-y\|_2
\end{align*}
\end{itemize}
\end{lemma}
\begin{proof}

{\bf Proof of Part 1.}
We can show
\begin{align*}
\| g_1(x) - g_1(y) \|_2
= & ~ \| f(x) \langle c(x), f(x) \rangle- f(y) \langle c(y), f(y) \rangle \|_2 \\
\leq & ~ \| f(x) \langle c(x), f(x) \rangle- f(y) \langle c(x), f(x) \rangle \|_2 \\
+ & ~ \| f(y) \langle c(x), f(x) \rangle- f(y) \langle c(y), f(x) \rangle \|_2 \\
+ & ~ \| f(y) \langle c(y), f(x) \rangle- f(y) \langle c(y), f(y) \rangle \|_2 
\end{align*}
where the 1st step follows from the definition of $g_1$,
the 2nd step follows from adding some terms $-f(y) \langle c(x), f(x) \rangle + f(y) \langle c(x), f(x) \rangle- f(y) \langle c(y), f(x) \rangle + f(y) \langle c(y), f(x) \rangle$ and $\|a + b\|_2 \leq \|a\|_2 + \|b\|_2$(Fact~\ref{fac:vector_norm}).

For the first term, we have
\begin{align*}
 \| f(x) \langle c(x), f(x) \rangle- f(y) \langle c(x), f(x) \rangle \|_2
 \leq & ~ \| f(x) - f(y) \|_2 \cdot | \langle c(x), f(x) \rangle | \\
 \leq & ~ \| f(x) - f(y) \|_2 \cdot \| c(x) \|_2 \cdot \| f(x) \|_2 \\
 \leq & ~ R_f \cdot \| x - y \|_2 \cdot \| c(x) \|_2 \cdot \| f(x) \|_2 
\end{align*}
where the 1st step follows from $\|\alpha a\|_2 \leq |\alpha| \|a\|_2$(Fact~\ref{fac:vector_norm}),
the 2nd step follows from $\langle a,b \rangle \leq \|a\|_2 \|b\|_2$(Fact~\ref{fac:vector_norm}),
the 3rd step follows from the definition of $R_f$.

For the second term, we have
\begin{align*}
\| f(y) \langle c(x), f(x) \rangle- f(y) \langle c(y), f(x) \rangle \|_2
\leq & ~ \| f(y) \|_2 \cdot | \langle c(x) - c(y) , f(x) \rangle | \\
\leq & ~ \| f(y) \|_2 \cdot \| c(x) - c(y) \|_2 \cdot \| f(x) \|_2 \\
\leq & ~ \| f(y) \|_2 \cdot R_f \cdot \| x - y \|_2 \cdot \| f(x) \|_2
\end{align*}
where the 1st step follows from $\|\alpha a\|_2 \leq |\alpha| \|a\|_2$(Fact~\ref{fac:vector_norm}),
the 2nd step follows from $\langle a,b \rangle \leq \|a\|_2 \|b\|_2$(Fact~\ref{fac:vector_norm}),
the 3rd step follows from the definition of $R_f$.

For the third term, we have
\begin{align*}
\| f(y) \langle c(y), f(x) \rangle- f(y) \langle c(y), f(y) \rangle \|_2 
\leq & ~ \| f(y) \|_2 \cdot | \langle c(y), f(x) - f(y) \rangle | \\
\leq & ~ \| f(y) \|_2 \cdot \| c(y) \|_2 \cdot \| f(x) - f(y) \|_2 \\
\leq & ~ \| f(y) \|_2 \cdot \| c(y) \|_2 \cdot R_f \cdot \| x - y \|_2
\end{align*}
the 2nd step follows from $\langle a,b \rangle \leq \|a\|_2 \|b\|_2$(Fact~\ref{fac:vector_norm}),
the 3rd step follows from the definition of $R_f$.

Combining three terms together, we complete the proof.

{\bf Proof of Part 2.}

We have
\begin{align*}
& ~ \| \diag( f(x) ) c(x) -  \diag( f(y) ) c(y) \|_2 \\
\leq & ~ \| \diag( f(x) ) c(x)  - \diag( f(x) ) c(y)  \|_2  + \| \diag( f(x) ) c(y) - \diag( f(y) ) c(y)\|_2 
\end{align*}
this step follows from adding terms $- \diag( f(x) ) c(y) + \diag( f(x) ) c(y)$.

For the first term, we have
\begin{align*}
\| \diag( f(x) ) c(x)  - \diag( f(x) ) c(y)  \|_2 \leq & ~ \| f(x) \|_{\infty} \cdot  \| c(x) - c(y) \|_2 \\
\leq & ~ \| f(x) \|_2 \cdot \| c(x) - c(y) \|_2 \\
\leq & ~ \| f(x) \|_2 \cdot R_f \cdot \| x - y \|_2
\end{align*}
where the 1st step follows from $\|Aa\| \leq \|A\|\|a\|_2$ where $A \in \R^{n \times d},b \in \R^{d \times 1}$(Fact~\ref{fac:matrix_norm}) and $\|\diag(a)\| \leq \|a\|_{\infty}$,
the 2nd step follows from $\|a\|_{\infty} \leq \|a\|_2$,
the 3rd step follows from the definition of $R_f$.

For the second term, we have
\begin{align*}
 \| \diag( f(x) ) c(y) - \diag( f(y) ) c(y)\|_2 
 \leq & ~ \| f(x) - f(y) \|_2 \cdot \| c(y) \|_2  \\
 \leq & ~ R_f \cdot \| x - y \|_2 \cdot  \| c(y) \|_2
\end{align*}
where the 1st step follows from $\diag(a) + \diag(b) = \diag(a + b)$, $\|Ab\| \leq \|A\| \|b\|_2$ where $A \in \R^{n \times d},b \in \R^{d \times 1}$ and $\|\diag(a)\| \leq \|a\|_{\infty} \leq \|a\|_2$,
the 2nd step follows from the definition of $R_f$.

Combining two terms together, then we complete the proof.
\end{proof}

{\bf Proof of Part 3.}

It follows from combining Part 1 and Part 2.

{\bf Proof of Part 4.}

It follows from Part 3.


\subsection{Hessian Calculations: Step 1, Hessian of \texorpdfstring{$\exp(Ax)$}{}}\label{sec:softmax:hessian_step_1}

\begin{lemma}[Hessian of $\exp(Ax)$]
If the following condition holds
\begin{itemize}
    \item Given a matrix $A \in \R^{n \times d}$.
\end{itemize}
Then, we have, for each $i \in [d]$
\begin{itemize}
    \item Part 1.
    \begin{align*}
        \frac{\d^2 \exp(Ax)}{\d x_i^2} = A_{*,i} \circ \exp(Ax) \circ A_{*,i}
    \end{align*}
    \item Part 2.
    \begin{align*}
        \frac{\d^2 \exp(Ax)}{\d x_i \d x_j} = A_{*,j} \circ \exp(Ax) \circ A_{*,i}
    \end{align*}

    \end{itemize}
\end{lemma}
\begin{proof}
    {\bf Proof of Part 1.}
    \begin{align*}
        \frac{ \d^2 ( \exp(Ax)  ) }{ \d x_i^2 }
            = & ~ \frac{\d}{\d x_i}(\frac{\d (\exp(Ax) )}{\d x_i} ) \\
            = & ~ \frac{\d (\exp(Ax) \circ A_{*,i})}{\d x_i} \\
            = & ~ A_{*,i} \circ \frac{\d \exp(Ax)}{\d x_i} \\
            = & ~ A_{*,i} \circ \exp(Ax) \circ A_{*,i}
    \end{align*}
    where the 1st step is an expansion of the Hessian, the 2nd step follows from the differential chain rule, the 3rd step extracts the matrix $A_{*,i}$ with constant entries out of the derivative, and the last step also follows from the chain rule.
    
    {\bf Proof of Part 2.}
    \begin{align*}
        \frac{ \d^2 ( \exp(Ax)  ) }{ \d x_i \d x_j }
            = & ~ \frac{\d}{\d x_i}(\frac{\d}{\d x_j}(\exp(Ax) ) ) \\
            = & ~ \frac{\d}{\d x_i}(\exp(Ax) \circ A_{*,j} ) \\
            = & ~ A_{*,j} \circ \exp(Ax) \circ A_{*,i}
    \end{align*}
    where the 1st step is an expansion of the Hessian, the 2nd and 3rd steps follow from the differential chain rule, the 3rd step follows from simple algebra.

\end{proof}

\newpage 
\subsection{Hessian Calculations: Step 2, Hessian of \texorpdfstring{$\alpha(x)$}{}}\label{sec:softmax:hessian_step_2}

\begin{lemma}
If the following conditions hold
\begin{itemize}
    \item Let $\alpha(x)$ be defined as Definition~\ref{def:alpha}.
\end{itemize}
Then, we have
\begin{itemize}
\item Part 1.
    \begin{align*}
        \frac{\d^2 \alpha(x) }{\d x_i^2} = \langle \exp(Ax), A_{*,i} \circ A_{*,i} \rangle
    \end{align*}
    \item Part 2.
    \begin{align*}
        \frac{\d^2 \alpha(x) }{\d x_i \d x_j} = \langle \exp(Ax),  A_{*,i} \circ A_{*,j} \rangle
    \end{align*}
\end{itemize}
\end{lemma}
\begin{proof}

{\bf Proof of Part 1.}
    \begin{align*}
        \frac{\d^2 \alpha(x) }{\d x_i^2}
        = & ~ \frac{\d}{\d x_i}(\frac{\d}{\d x_i} \langle \exp(Ax) , {\bf 1}_n \rangle) \\
        = & ~ \frac{\d}{\d x_i}(\langle \exp(Ax) \circ A_{*,i}, {\bf 1}_n \rangle) \\
        = & ~ \langle A_{*,i} \circ \exp(Ax) \circ A_{*,i}, {\bf 1}_n \rangle \\
        = & ~ \langle \exp(Ax) , A_{*,i} \circ A_{*,i} \rangle 
    \end{align*} 
    where the 1st step follows from the expansion of hessian, the 2nd step follows from Part 3 of Lemma~\ref{lem:gradient_computation}, the 3rd step follows from simple algebra, and the last step follows from Fact~\ref{fac:basic}.

    {\bf Proof of Part 2.}
    \begin{align*}
        \frac{\d^2 \alpha(x) }{\d x_i \d x_j}
        = & ~ \frac{\d}{\d x_j}(\frac{\d}{\d x_i} \langle \exp(Ax) , {\bf 1}_n \rangle) \\
        = & ~ \frac{\d}{\d x_j}(\langle \exp(Ax) \circ A_{*,i}, {\bf 1}_n \rangle) \\
        = & ~ \langle A_{*,j} \circ \exp(Ax) \circ A_{*,i}, {\bf 1}_n \rangle \\
        = & ~ \langle \exp(Ax) , A_{*,i} \circ A_{*,j}\rangle 
    \end{align*} 
    where the 1st step follows from the expansion of hessian, the 2nd step follows from Part 2 of Lemma~\ref{lem:gradient_computation}, the 3rd step follows from simple algebra, the last step follows from Fact~\ref{fac:basic}.

\end{proof}


\subsection{Hessian Calculations: Step 3, Hessian of \texorpdfstring{$\alpha(x)^{-1}$}{}}\label{sec:softmax:hessian_step_3}

\begin{lemma}[Hessian of $\alpha(x)^{-1}$]
If the following conditions hold
\begin{itemize}
    \item Let $\alpha(x)$ be defined as Definition~\ref{def:alpha}
    \item Let $f(x)$ be defined in Definition~\ref{def:f}.
\end{itemize}
We have
\begin{itemize}
    \item Part 1. 
    \begin{align*}
        \frac{\d^2 \alpha(x)^{-1}}{\d x_i^2} = & ~ 2 \alpha(x)^{-1} \cdot \langle f(x), A_{*,i} \rangle^2 - \alpha(x)^{-1} \cdot \langle f(x), A_{*,i} \circ A_{*,i} \rangle \\
        = & ~ 2 \alpha(x)^{-1} A_{*,i}^\top f(x) f(x)^\top A_{*,i}   -  A_{*,i}^\top \diag(f(x)) A_{*,i}
    \end{align*}
    \item Part 2.
    \begin{align*}
        \frac{\d^2 \alpha(x)^{-1}}{\d x_i \d x_j} 
        = & ~  2 \alpha(x)^{-1} \langle f(x), A_{*,i} \rangle \langle f(x), A_{*,j} \rangle  - \alpha(x)^{-1} \langle f(x), A_{*,i} \circ A_{*,j} \rangle \\
        = & ~ 2 \alpha(x)^{-1} A_{*,i}^\top f(x) f(x)^\top A_{*,j}   -  A_{*,i}^\top \diag(f(x)) A_{*,j}
    \end{align*}
\end{itemize}
\end{lemma}
\begin{proof}
{\bf Proof of Part 1.}
    \begin{align*}
         \frac{\d^2 \alpha(x)^{-1}}{\d x_i^2}
        = & ~ \frac{\d}{\d x_i}( \frac{\d}{ \d x_i} \alpha(x)^{-1}) \\
        = & ~ \frac{\d}{\d x_i}( - \alpha(x)^{-1} \langle f(x), A_{*,i} \rangle ) \\
        = & ~ - ( \frac{\d}{\d x_i} \alpha(x)^{-1} ) \cdot  \langle f(x), A_{*,i} \rangle - \alpha(x)^{-1} \frac{\d}{\d x_i} \langle f(x), A_{*,i} \rangle \\
        = & ~ 2 \alpha(x)^{-1} \langle f(x), A_{*,i} \rangle^2 - \alpha(x)^{-1} \langle f(x), A_{*,i} \circ A_{*,i} \rangle
    \end{align*}
    where the 1st step follows from the expansion of hessian, the 2nd step follows from Part 3 of Lemma~\ref{lem:gradient_computation}, the 3rd step follows from differential chain rule, the 4th step follows from simple algebra, the last step follows from Fact~\ref{fac:basic}.

    {\bf Proof of Part 2.}
    \begin{align*}
        \frac{\d^2 \alpha(x)^{-1}}{\d x_i \d x_j}
        = & ~ \frac{\d}{\d x_j}(\frac{\d}{\d x_i} \alpha(x)^{-1}) \\
        = & ~ \frac{\d}{\d x_j}( -\alpha(x)^{-1} \langle f(x), A_{*,i} \rangle ) \\
        = & ~ - ( \frac{\d}{\d x_j} \alpha(x)^{-1} ) \cdot  \langle f(x), A_{*,i} \rangle - \alpha(x)^{-1} \frac{\d}{\d x_j} \langle f(x), A_{*,i} \rangle \\
        = & ~ 2 \alpha(x)^{-1} \langle f(x), A_{*,i} \rangle \langle f(x), A_{*,j} \rangle -  \alpha(x)^{-1} \langle f(x), A_{*,j} \circ A_{*,i} \rangle
    \end{align*}

    where the 1st step follows from the expansion of hessian, the 2nd step follows from Part 3 of Lemma~\ref{lem:gradient_computation}, the 3rd step follows from differential chain rule, the 4th step follows from basic differential rule, the 5th step step follows from simple algebra, the last step follows from Fact~\ref{fac:basic}.
\end{proof}


\subsection{Hessian Calculations: Step 4, Hessian of \texorpdfstring{$f(x)$}{}}\label{sec:softmax:hessian_step_4}

\begin{lemma}[Hessian of $f(x)$]
 If the following conditions hold
 \begin{itemize}
 \item Let $f(x) = \langle \exp(Ax) , {\bf 1}_n \rangle^{-1} \exp(Ax)$ (see Definition~\ref{def:f}).
\end{itemize}

Then, we have
\begin{itemize}
\item Part 1.
    \begin{align*}
        \frac{\d^2 f(x) }{\d x_i^2}
        = & ~ ~ 2  \langle f(x), A_{*,i} \rangle^2 \cdot f(x)   - \langle f(x) , A_{*,i} \circ A_{*,i} \rangle \cdot f(x) \\
        & ~ -2  \langle f(x), A_{*,i} \rangle f(x) \circ A_{*,i}   + A_{*,i} \circ f(x) \circ A_{*,i}
    \end{align*}
    \item Part 2.
    \begin{align*}
        \frac{\d^2 f(x) } {\d x_i \d x_j}
        = & ~ 2  \langle f(x), A_{*,i} \rangle \langle f(x), A_{*,j} \rangle f(x)   -   \langle f(x) , A_{*,i} \circ A_{*,j} \rangle f(x) \\
        & ~ -  \langle f(x) , A_{*,i} \rangle f(x) \circ A_{*,j} 
         -   \langle f(x), A_{*,j} \rangle f(x) \circ A_{*,i}   + A_{*,i} \circ f(x) \circ A_{*,j}
    \end{align*}
\end{itemize}
\end{lemma}
\begin{proof}
{\bf Proof of Part 1.}
    \begin{align*}
         \frac{\d^2 f(x) }{\d x_i^2} 
        = & ~ \frac{\d}{\d x_i}(\frac{\d}{\d x_i} f(x) ) \\
        = & ~ \frac{\d }{\d x_i} ( - \langle f(x), A_{*,i} \rangle \cdot f(x) + f(x) \circ A_{*,i} ) \\
        = & ~ 2  \langle f(x), A_{*,i} \rangle^2 \cdot f(x)   - \langle f(x) , A_{*,i} \circ A_{*,i} \rangle \cdot f(x) \\
        & ~ -2  \langle f(x), A_{*,i} \rangle f(x) \circ A_{*,i}   + A_{*,i} \circ f(x) \circ A_{*,i}
    \end{align*}
    where the 1st step follows from the expansion of hessian, the 2nd step follows from Part 4 of Lemma~\ref{lem:gradient_computation} and differential chain rule.

    {\bf Proof of Part 2.}
    \begin{align*}
        & \frac{\d^2 f(x) }{\d x_i \d x_j} \\
        = & ~ \frac{\d}{\d x_j}(\frac{\d}{\d x_i} f(x) ) \\
        = & ~ \frac{\d}{\d x_j}(   - \langle f(x), A_{*,i} \rangle \cdot f(x) + f(x) \circ A_{*,i} ) \\
        = & ~ 2  \langle f(x), A_{*,i} \rangle \langle f(x), A_{*,j} \rangle f(x)   -   \langle f(x) , A_{*,i} \circ A_{*,j} \rangle f(x) \\
        & ~ -  \langle f(x) , A_{*,i} \rangle f(x) \circ A_{*,j} 
         -   \langle f(x), A_{*,j} \rangle f(x) \circ A_{*,i}   + A_{*,i} \circ f(x) \circ A_{*,j}
    \end{align*}
    where the 1st step follows from the expansion of hessian, the 2nd step follows from Part 4 of Lemma~\ref{lem:gradient_computation}, the 3rd step follows from differential chain rule.

\end{proof}


\subsection{Hessian Calculations: Step 5, Hessian of \texorpdfstring{$L_{\exp}(x)$}{}}\label{sec:softmax:hessian_step_5}
 
\begin{lemma}[Hessian of $L_{\exp}(x)$]\label{lem:hessian_of_L_x}
We define
\begin{itemize}
    \item $B_1(x) \in \R^{n \times n}$ such that 
    \begin{align*}
        A_{*,i}^\top B_1(x) A_{*,j} := (-   \langle f(x), A_{*,j} \rangle f(x) + f(x) \circ A_{*,j})^\top   \cdot (-  \langle f(x), A_{*,i} \rangle f(x) ) +    f(x) \circ A_{*,i}))
    \end{align*}
    \item $B_2(x) \in \R^{n \times n}$ such that
    \begin{align*}
        A_{*,i}^\top B_2(x) A_{*,j} := & ~ c^\top \cdot (2 \langle f(x), A_{*,i} \rangle \langle f(x), A_{*,j} \rangle f(x)  -   \langle f(x) , A_{*,i} \circ A_{*,j} \rangle  f(x)  \\ 
        & ~ -  \langle f(x), A_{*,i} \rangle f(x) \circ A_{*,j}  
          -  \langle f(x), A_{*,j} \rangle f(x) \circ A_{*,i} +   A_{*,i} \circ f(x) \circ A_{*,j})
    \end{align*}
\end{itemize}
Then we have 
\begin{itemize}
\item Part 1.
    \begin{align*}
         \frac{\d^2 L}{\d x_i^2} 
        = A_{*,i}^\top B_1(x) A_{*,i} + A_{*,i}^\top B_2(x) A_{*,i}
    \end{align*}
    \item Part 2.
    \begin{align*}
         \frac{\d^2 L}{\d x_i \d x_j} 
        = A_{*,i}^\top B_1(x) A_{*,j} + A_{*,i}^\top B_2(x) A_{*,j}
    \end{align*}
\end{itemize}
\end{lemma}

\begin{proof}
 {\bf Proof of Part 1.}
    \begin{align*}
        & \frac{\d^2 L}{\d x_i^2} = \frac{\d}{\d x_i}(\frac{\d L}{\d x_i}) \\
        = & ~ \frac{\d}{\d x_i }((f(x) - b)^\top(-\langle f(x) \circ A_{*,i}, {\bf 1}_n \rangle f(x)) + f(x) \circ A_{*,i}) \\
        = & ~ (- \langle f(x), A_{*,i} \rangle f(x) + f(x) \circ A_{*,i})^\top) \cdot (-\langle f(x), A_{*,i} \rangle f(x)) +  f(x) \circ A_{*,i})) \\
         & ~ + c^\top\cdot (2 \langle f(x), A_{*,i} \rangle^2 f(x) - \langle f(x) , A_{*,i} \circ A_{*,j} \rangle  f(x)-2\langle f(x), A_{*,i} \rangle f(x) \circ A_{*,i} + A_{*,i} \circ f(x) \circ A_{*,i}) \\
         = & ~ A_{*,i}^\top B_1(x) A_{*,i} + A_{*,i}^\top B_2(x) A_{*,i}
    \end{align*}
    where the 1st step follows from the expansion of hessian, the 2nd step follows from differential chain rule.

    {\bf Proof of Part 2.}
    \begin{align*}
        & \frac{\d^2 L}{\d x_i \d x_j} = \frac{\d}{\d x_j}(\frac{\d L}{\d x_i}) \\
        = & ~ \frac{\d}{\d x_j}(( f(x) - b)^\top(-\langle f(x), A_{*,i} \rangle f(x))+  f(x) \circ A_{*,i}) \\
        = & (-   \langle f(x), A_{*,j} \rangle f(x) + f(x) \circ A_{*,j})^\top   \cdot (-  \langle f(x), A_{*,i} \rangle f(x) ) +    f(x) \circ A_{*,i})) \\
         & ~ + c^\top \\
         & ~ \cdot (2 \langle f(x), A_{*,i} \rangle \langle f(x), A_{*,j} \rangle f(x)  -   \langle f(x) , A_{*,i} \circ A_{*,j} \rangle  f(x)  -  \langle f(x), A_{*,i} \rangle f(x) \circ A_{*,j}  \\
         & ~ -  \langle f(x), A_{*,j} \rangle f(x) \circ A_{*,i} +   A_{*,i} \circ f(x) \circ A_{*,j}) \\
         = & ~ A_{*,i}^\top B_1(x) A_{*,j} + A_{*,i}^\top B_2(x) A_{*,j}
    \end{align*}
    where the 1st step follows from the expansion of hessian, the 2nd step follows from differential chain rule, the 3rd step is a simplification of step 2 by applying notations $\alpha$ (Definition~\ref{def:alpha}) and $c$ ( Definition~\ref{def:c} ).

\end{proof}


\subsection{Helpful Lemma}\label{sec:softmax:help}
 
The goal of this section to prove Lemma~\ref{lem:key_lemma}. We remark that in this lemma, we can replace $f(x)$ by any vector. However, for easy of presentation, we use $f(x)$.
\begin{lemma}\label{lem:key_lemma}
For any length-$n$ vector $c \in \R^n$ and any vector $f(x) \in \R^n$, 
we have
\begin{itemize}
    \item Part 1.
    \begin{align*}
        c^\top ( A_{*,i} \circ f(x) \circ A_{*,j} ) = A_{*,i}^\top \underbrace{ \diag ( c \circ f(x) ) }_{n \times n} A_{*,j}
    \end{align*}
    \item Part 2.
    \begin{align*}
        c^\top f(x) \langle f(x), A_{*,i} \rangle \langle f(x), A_{*,j} \rangle = A_{*,i}^\top \underbrace{ f(x) }_{n \times 1} \underbrace{ \langle c, f(x) \rangle }_{ \mathrm{scalar} } \underbrace{ f(x)^\top }_{1 \times n} A_{*,j}
    \end{align*}
    \item Part 3.
    \begin{align*}
        c^\top \langle f(x), A_{*,i} \circ A_{*,j} \rangle f(x) = A_{*,i}^\top \underbrace{ \diag( \langle c, f(x) \rangle  f(x) ) }_{n \times n}  A_{*,j}
    \end{align*}
    \item Part 4. 
    \begin{align*}
        c^\top \langle f(x), A_{*,j} \rangle f(x) \circ A_{*,i} = A_{*,i}^\top \underbrace{ (c \circ f(x) ) }_{n \times 1} \underbrace{ f(x)^\top }_{1 \times n} A_{*,j}
    \end{align*}
    \item Part 5. 
    \begin{align*}
        c^\top \langle f(x), A_{*,i} \rangle f(x) \circ A_{*,j} = A_{*,i}^\top \underbrace{ f(x) }_{n \times 1} \underbrace{( f(x) \circ c)^\top }_{1 \times n} A_{*,j}
    \end{align*}
    \item Part 6.
    \begin{align*}
        ( \langle f(x), A_{*,j} \rangle f(x) )^\top f(x) \circ A_{*,i} = & ~ A_{*,i} \underbrace{ ( f(x) \circ f(x) ) }_{n \times 1} \underbrace{ f(x)^\top }_{1 \times n} A_{*,j} 
    \end{align*}
    \item Part 7.
    \begin{align*}
        (f(x) \circ A_{*,i} )^\top ( f(x) \circ A_{*,j} ) = A_{*,i}^\top \underbrace{ \diag( f(x) \circ f(x) ) }_{n \times n} A_{*,j}
    \end{align*}
    \item Part 8. 
    \begin{align*}
        ( \langle f(x), A_{*,j} \rangle f(x) )^\top ( \langle f(x), A_{*,i} \rangle f(x) )
        = & ~ A_{*,i}^\top \underbrace{ f(x) }_{n \times 1} \underbrace{ \langle f(x), f(x) \rangle }_{\mathrm{scalar}} \underbrace{ f(x)^\top }_{1 \times n} A_{*,j} 
    \end{align*}
    \item Part 9.
    \begin{align*}
        (f(x) \circ A_{*,i} )^\top ( f(x) \circ A_{*,j} ) = A_{*,i}^\top \diag(f(x) \circ f(x)) A_{*,j} 
    \end{align*}
\end{itemize}
\end{lemma}
 
\begin{proof}

    {\bf Proof of Part 1.}
    \begin{align*}
        c^\top ( A_{*,i} \circ f(x) \circ A_{*,j} )
        = & ~ A_{*,i}^\top (c \circ f(x) \circ A_{*,j}) \\
        = & ~ A_{*,i}^\top \diag(c \circ f(x)) \circ A_{*,j}
    \end{align*}
    where the 1st step follows from Fact~\ref{fac:circ_diag}, 
    the 2nd step follows from Fact~\ref{fac:circ_diag}.

    {\bf Proof of Part 2.}
    \begin{align*}
        c^\top f(x) \langle f(x), A_{*,i} \rangle \langle f(x), A_{*,j} \rangle
        = & ~ \langle c, f(x) \rangle \langle f(x), A_{*,i} \rangle \langle f(x), A_{*,j} \rangle \\
        = & ~ A_{*,i}^\top f(x) \langle c, f(x) \rangle (f(x))^\top A_{*,j}
    \end{align*}
    where the 1st step follows from $a^\top b = \langle a, b \rangle$ (Fact~\ref{fac:basic}), 
    the 2nd step follows from $\langle a,b \rangle = a^\top b$ (Fact~\ref{fac:basic}).

    {\bf Proof of Part 3.}
    \begin{align*}
        c^\top \langle f(x), A_{*,i} \circ A_{*,j} \rangle f(x)
        = & ~ c^\top (f(x))^\top A_{*,i} \circ A_{*,j} f(x) \\
        = & ~ A_{*,i}^\top (f(x))^\top c \circ A_{*,j} f(x) \\
        = & ~ A_{*,i}^\top \langle f(x), c \rangle \circ A_{*,j} f(x) \\
        = & ~ A_{*,i}^\top \diag(\langle f(x), c \rangle)f(x) A_{*,j}
    \end{align*}
    where the 1st step follows from $\langle a, b \rangle = a^\top b$ (Fact~\ref{fac:basic}), 
    the 2nd step follows from Fact~\ref{fac:circ_diag}, 
    the 3rd step follows from $a^\top b = \langle a,b \rangle$ (Fact~\ref{fac:basic}), 
    the last step follows from Fact~\ref{fac:circ_diag}.

    {\bf Proof of Part 4.}
    \begin{align*}
        c^\top \langle f(x), A_{*,j} \rangle f(x) \circ A_{*,i}
        = & ~ c^\top (f(x))^\top A_{*,j} f(x) \circ A_{*,i} \\
        = & ~ A_{*,i}^\top (f(x))^\top A_{*,j} f(x) \circ c \\
        = & ~ A_{*,i}^\top (f(x) \circ c) (f(x))^\top A_{*,j}
    \end{align*}
    where the 1st step follows from $\langle a,b \rangle = a^\top b$ (Fact~\ref{fac:basic}), 
    the 2nd step follows from Fact~\ref{fac:circ_diag}, 
    the 3rd step follows from $f(x)^\top A_{*,j} = \langle f(x), A_{*,j} \rangle$ (Fact~\ref{fac:basic}) is a scalar. 

    {\bf Proof of Part 5.}
    \begin{align*}
        c^\top \langle f(x), A_{*,i} \rangle f(x) \circ A_{*,j}
        = & ~ (f(x))^\top A_{*,i} c^\top f(x) \circ A_{*,j} \\
        = & ~ (f(x))^\top A_{*,i} A_{*,j}^\top f(x) \circ c \\
        = & ~ (f(x))^\top A_{*,i} (f(x) \circ c)^\top A_{*,j} \\
        = & ~ A_{*,i}^\top f(x) (f(x) \circ c)^\top A_{*,j}
    \end{align*}
    where the 1st step follows from $\langle a,b \rangle = a^\top b$ (Fact~\ref{fac:basic}), 
    the 2nd step follows from Fact~\ref{fac:circ_diag}, 
    the 3rd step follows from $a^\top b = b^\top a$ (Fact~\ref{fac:basic}), 
    the last step follows from $a^\top b = b ^\top a$ (Fact~\ref{fac:basic}).

    {\bf Proof of Part 6}
    \begin{align*}
        ( \langle f(x), A_{*,j} \rangle f(x) )^\top f(x) \circ A_{*,i}
        = & ~ A_{*,i}^\top f(x) \circ \langle f(x), A_{*,j} \rangle f(x) \\
        = & ~ A_{*,i}^\top f(x) \circ (f(x))^\top A_{*,j} f(x) \\
        = & ~ A_{*,i}^\top f(x) \circ f(x) (f(x))^\top A_{*,j}
    \end{align*}
    where the 1st step follows from Fact~\ref{fac:circ_diag}, 
    the 2nd step follows from $\langle a,b \rangle = a^\top b$ (Fact~\ref{fac:basic}), 
    the 3rd step follows from $f(x)^\top A_{*,j} = \langle f(x), A_{*,j} \rangle$ is a scalar (Fact~\ref{fac:basic}).

    {\bf Proof of Part 7.}
    \begin{align*}
        (f(x) \circ A_{*,i} )^\top ( f(x) \circ A_{*,j} ) 
        = & ~ \langle f(x) \circ A_{*,i}, f(x) \circ A_{*,j} \rangle \\
        = & ~ \langle f(x) \circ f(x), A_{*,i} \circ A_{*,j} \rangle \\
        = & ~ (f(x) \circ f(x))^\top (A_{*,i} \circ A_{*,j}) \\
        = & ~ A_{*,i}^\top (f(x) \circ f(x) \circ A_{*,j}) \\
        = & ~ A_{*,i}^\top \diag(f(x) \circ f(x)) A_{*,j}
    \end{align*}
    where the 1st step follows from $a^\top b = \langle a,b \rangle$ (Fact~\ref{fac:basic}), 
    the 2nd step follows from Fact~\ref{fac:basic}, 
    the 3rd step follows from $\langle a,b \rangle = a^\top b$ (Fact~\ref{fac:basic}), 
    the 4th step follows from Fact~\ref{fac:circ_diag}, 
    the last step follows from Fact~\ref{fac:circ_diag}.

    {\bf Proof of Part 8.}

    \begin{align*}
        ( \langle f(x), A_{*,j} \rangle f(x) )^\top ( \langle f(x), A_{*,i} \rangle f(x) )
         = & ~ \langle f(x), A_{*,j} \rangle f(x)^\top ( \langle f(x), A_{*,i} \rangle f(x) )\\
         = & ~ f(x)^\top A_{*,j} f(x)^\top f(x)^\top A_{*,i} f(x) \\
         = & ~ f(x)^\top A_{*,i} f(x)^\top A_{*,j} f(x)^\top f(x) \\
         = & ~ A_{*,i}^\top f(x) f(x)^\top A_{*,j} f(x)^\top f(x) \\
         = & ~ A_{*,i}^\top f(x) f(x)^\top f(x) f(x)^\top A_{*,j} \\
         = & ~ A_{*,i}^\top f(x)\langle f(x), f(x) \rangle f(x)^\top A_{*,j}
    \end{align*}
    where the 1st step follows from $a^\top b = b^\top a$ (Fact~\ref{fac:basic}), 
    the 2nd step follows from $\langle a,b \rangle = a^\top b$ (Fact~\ref{fac:basic}), 
    the 3rd step follows from $a^\top b = b^\top a$ (Fact~\ref{fac:basic}), 
    the 4th step follows from $A_{*,i}^\top f(x) = \langle A_{*,i},f(x) \rangle$ is a scalar (Fact~\ref{fac:basic}),
    the 5th step step follows from $f(x)^\top A_{*,j} = \langle f(x), A_{*,j} \rangle$ is a scalar (Fact~\ref{fac:basic}),
    the last step follows from $a^\top b = \langle a,b \rangle$ (Fact~\ref{fac:basic}).

     {\bf Proof of Part 9.}
     \begin{align*}
        (f(x) \circ A_{*,i} )^\top ( f(x) \circ A_{*,j} ) 
        = & ~ \langle f(x) \circ A_{*,i}, f(x) \circ A_{*,j} \rangle \\
        = & ~ \langle f(x) \circ f(x), A_{*,i} \circ A_{*,j} \rangle \\
        = & ~ (f(x) \circ f(x))^\top (A_{*,i} \circ A_{*,j}) \\
        = & ~ A_{*,i}^\top (f(x) \circ f(x) \circ A_{*,j}) \\
        = & ~ A_{*,i}^\top \diag(f(x) \circ f(x)) A_{*,j}
     \end{align*}
     where the 1st step follows from $a^\top b = \langle a,b \rangle$ (Fact~\ref{fac:basic}),
     the 2nd step follows from Fact~\ref{fac:basic} , 
     the 3rd step follows from $\langle a,b \rangle = a^\top b$ (Fact~\ref{fac:basic}),
     the 4th step follows from Fact~\ref{fac:circ_diag},
     the last step follows from Fact~\ref{fac:circ_diag}.
\end{proof}

\subsection{Decomposing \texorpdfstring{$B_1(x)$}{},  \texorpdfstring{$B_2(x)$}{} and \texorpdfstring{$B(x)$}{} into Low Rank Plus Diagonal}\label{sec:softmax:decompose}

\begin{lemma}[Rewriting $B_1(x)$ and $B_2(x)$]\label{lem:B_1_B_2_B}
If the following conditions hold
\begin{itemize}
    \item Given matrix $A \in \R^{n \times d}$.
    \item Let $f(x)$ be defined as Definition~\ref{def:f}.
    \item Let $c(x)$ be defined as Definition~\ref{def:c}.
    \item Let $B(x) = B_1(x) + B_2(x)$.
\end{itemize}
Then, we can show that
\begin{itemize}
    \item {\bf Part 1.} For $B_1(x) \in \R^{n \times n}$, we have
    \begin{align*}
        B_1(x) = & ~ \underbrace{ \langle f(x),f(x) \rangle }_{ \mathrm{scalar} } \cdot \underbrace{ f(x) }_{n \times 1} \underbrace{ f(x)^\top }_{1 \times n} + \underbrace{ \diag( f(x) \circ f(x) ) }_{n \times n~\mathrm{diagonal~matrix}} \\
        & ~ + \underbrace{ (f(x) \circ f(x)) }_{n \times 1} \cdot \underbrace{ f(x)^\top }_{1 \times n} + \underbrace{ (f(x) \circ f(x)) }_{n \times 1} \cdot \underbrace{ f(x)^\top }_{1 \times n}
    \end{align*}
    \begin{itemize}
        \item In summary, $B_1(x) \in \R^{n \times n}$ is constructed by three rank-$1$ matrices and a diagonal matrix.
    \end{itemize}
    \item {\bf Part 2.} For $B_2(x) \in \R^{n \times n}$, we have
    \begin{align*}
        B_2(x) = & ~ \underbrace{ 2 \langle c(x),f(x) \rangle }_{\mathrm{scalar}} \cdot \underbrace{ f(x) }_{n \times 1}  \underbrace{ f(x)^\top }_{1 \times n} + \underbrace{ \langle c(x),f(x) \rangle }_{\mathrm{scalar}} \cdot \underbrace{ \diag( f(x)) }_{n \times n \mathrm{~diagonal~matrix} }  + \underbrace{ \diag(c(x) \circ f(x)) }_{ n \times n \mathrm{~diagonal~matrix} } \\
        & ~ - \underbrace{ (c(x) \circ f(x)) }_{n \times 1} \underbrace{ f(x)^\top }_{1 \times n} - \underbrace{ f(x) }_{n \times 1} \underbrace{ (f(x) \circ c(x) )^\top }_{1 \times n}
    \end{align*}
    \begin{itemize}
        \item In summary, $B_2(x) \in \R^{n \times n}$ is constructed by three rank-$1$ matrices and two diagonal matrices.
    \end{itemize}
    \item {\bf Part 3.} For $B(x) \in \R^{n \times n}$, we have
    \begin{align*}
        B(x) = & ~ \underbrace{ \langle 3 f(x) - 2 b, f(x) \rangle }_{ \mathrm{scalar} } \cdot \underbrace{ f(x) }_{n \times 1}  \underbrace{ f(x)^\top }_{1 \times n} \\
        & ~ + \underbrace{ \langle f(x) - b , f(x) \rangle }_{ \mathrm{scalar} } \cdot \underbrace{ \diag(f(x)) }_{n \times n \mathrm{~diagonal~matrix} } \\
        & ~ + \underbrace{ \diag( ( 2 f(x) - b) \circ f(x) ) }_{n \times n \mathrm{~diagonal~matrix}} \\
        & ~ + \underbrace{ (b \circ f(x)) }_{n \times 1} \cdot \underbrace{ f(x)^\top }_{1 \times n} + \underbrace{ f(x) }_{n \times 1} \cdot \underbrace{ (b \circ f(x))^\top }_{1 \times n}
    \end{align*}
    \begin{itemize}
        \item In summary, $B(x) \in \R^{n \times n}$ is constructed by three rank-$1$ matrices and two diagonal matrices.
    \end{itemize}
\end{itemize}
\end{lemma}

\begin{proof}
{\bf Proof of Part 1. $B_1(x)$.}

    For $B_1(x)$, we have:
    \begin{align}\label{eq:hessian_simplification_step_1}
        A_{*,i}^\top B_1(x) A_{*,j}
        = & ~ (-   \langle f(x), A_{*,j} \rangle f(x) + f(x) \circ A_{*,j})^\top   \cdot (-  \langle f(x), A_{*,i} \rangle f(x) ) +    f(x) \circ A_{*,i})) \notag \\
        = & ~ (-(\langle f(x), A_{*,j} \rangle f(x))^\top + (f(x) \circ A_{*,j})^\top) \cdot (-  \langle f(x), A_{*,i} \rangle f(x) ) +    f(x) \circ A_{*,i})) \notag \\
        = & ~ (\langle f(x), A_{*,j} \rangle f(x))^\top\langle f(x), A_{*,i} \rangle f(x) + (f(x) \circ A_{*,j})^\top (f(x) \circ A_{*,i})) \notag \\
        & ~ -(\langle f(x), A_{*,j} \rangle f(x))^\top (f(x) \circ A_{*,i}) - (f(x) \circ A_{*,j})^\top \langle f(x), A_{*,i} \rangle f(x) \notag \\
        = & ~ A_{*,i}^\top f(x) \langle f(x),f(x) \rangle f(x)^\top A_{*,j} + A_{*,i}^\top \diag(f(x) \circ f(x))A_{*,j} \notag \\
        & ~ - A_{*,i}^\top (f(x) \circ f(x)) f(x)^\top A_{*,j} - A_{*,i}^\top (f(x) \circ f(x))^\top f(x) A_{*,j}
    \end{align}
    where the 1st step follows from the definition of $B_1(x)$, the 2nd step follows from $(A+B)^\top=A^\top+B^\top$, the 3rd step follows from simple algebra, the last step follows from Lemma~\ref{lem:key_lemma}.
    
Thus, by extracting $A_{*,i}^\top$ and $A_{*,j}$, we have:
\begin{align*}
 B_1(x) = & ~ \langle f(x),f(x) \rangle \cdot f(x) f(x)^\top + \diag(f(x) \circ f(x)) \\
 & ~ +  (f(x) \circ f(x)) f(x)^\top + f(x)(f(x) \circ f(x))^\top
\end{align*}

{\bf Proof of Part 2. $B_2(x)$.}
    
    For $B_2(x) \in \R^{n \times n}$, we have:
    \begin{align*}
       & ~ A_{*,i}^\top B_2(x) A_{*,j} \\
        = & ~ c(x)^\top \cdot (2 \langle f(x), A_{*,i} \rangle \langle f(x), A_{*,j} \rangle f(x)  -   \langle f(x) , A_{*,i} \circ A_{*,j} \rangle  f(x)  -  \langle f(x), A_{*,i} \rangle f(x) \circ A_{*,j}  \\
        & ~ -  \langle f(x), A_{*,j} \rangle f(x) \circ A_{*,i} +   A_{*,i} \circ f(x) \circ A_{*,j})
    \end{align*}
    
    Thus, we can rewrite $B_2(x)$ as
    \begin{align}\label{eq:hessian_simplification_step_2}
        & ~ A_{*,i}^\top B_2(x) A_{*,j} \notag \\
        = & ~ c(x)^\top \cdot (2 \langle f(x), A_{*,i} \rangle \langle f(x), A_{*,j} \rangle f(x)  -   \langle f(x) , A_{*,i} \circ A_{*,j} \rangle  f(x)  -  \langle f(x), A_{*,i} \rangle f(x) \circ A_{*,j} \notag  \\
        & ~ -  \langle f(x), A_{*,j} \rangle f(x) \circ A_{*,i} +   A_{*,i} \circ f(x) \circ A_{*,j}) \notag \\
        = & ~ 2c(x)^\top\langle f(x), A_{*,i} \rangle \langle f(x), A_{*,j} \rangle f(x) - c(x)^\top \langle f(x) , A_{*,i} \circ A_{*,j} \rangle  f(x) + c(x)^\top A_{*,i} \circ f(x) \circ A_{*,j} \notag \\
        & ~ - c(x)^\top \langle f(x), A_{*,i} \rangle f(x) \circ A_{*,j} - c(x)^\top \langle f(x), A_{*,j} \rangle f(x) \circ A_{*,i} \notag \\
        = & ~ 2A_{*,i}^\top f(x) \langle c(x),f(x) \rangle f(x)^\top A_{*,j} - A_{*,i}^\top \diag(\langle c(x),f(x) \rangle f(x)) A_{*,j} + A_{*,i}^\top \diag(c(x) \circ f(x)) A_{*,j} 
 \notag\\
        & ~ - A_{*,i}^\top f(x) (f(x) \circ c)^\top A_{*,j} - A_{*,i} (c(x) \circ f(x)) f(x)^\top A_{*,j}
    \end{align}
    where the 1st step follows from definition of $B_2(x)$, the 2nd step follows from simple algebra, the 3rd step follows from simple algebra, the last step follows from Lemma~\ref{lem:key_lemma}.
    
By extracting $A_{*,i}^\top$ and $A_{*,j}$, we have
\begin{align*}
B_2(x) = & ~ 2 \langle c,f(x) \rangle f(x) f(x)^\top + \diag(\langle c,f(x) \rangle f(x)) + \diag(c \circ f(x))\\
& ~ -  (c (x) \circ f(x)) f(x)^\top - f(x) ( c(x) \circ f(x) )^\top 
\end{align*}
    
{\bf Proof of Part 3. $B(x)$}

We define 
\begin{align*}
    B_{1,1}(x) := & ~ \langle f(x),f(x) \rangle \cdot f(x) f(x)^\top \\
    B_{1,2}(x) := & ~ \diag(f(x) \circ f(x)) \\
    B_{1,3}(x) := & ~ (f(x) \circ f(x)) f(x)^\top \\
    B_{1,4}(x) := & ~ f(x)(f(x) \circ f(x))^\top
\end{align*}

Thus, we have:
\begin{align*}
    B_1(x) = B_{1,1}(x) + B_{1,2}(x) + B_{1,3}(x) + B_{1,4}(x)
\end{align*}

Similarly, we define
\begin{align*}
    B_{2,1}(x) := & ~ 2 \langle c,f(x) \rangle f(x) f(x)^\top \\
    B_{2,2}(x) := & ~ \diag(\langle c,f(x) \rangle f(x)) \\
    B_{2,3}(x) := & ~ \diag(c \circ f(x)) \\
    B_{2,4}(x) := & ~ -  (c (x) \circ f(x)) f(x)^\top \\
    B_{2,5}(x) := & ~ - f(x) ( c(x) \circ f(x) )^\top 
\end{align*}

Thus, we have:
\begin{align*}
    B_2(x) = B_{2,1}(x) + B_{2,2}(x) + B_{2,3}(x) + B_{2,4}(x) + B_{2,5}(x)
\end{align*}

Merge $B_{1,1}(x)$ and $B_{2,1}(x)$:
\begin{align*} 
    B_{1,1}(x) + B_{2,1}(x)
   = & ~ \langle f(x),f(x) \rangle \cdot f(x) f(x)^\top + 2 \langle c(x),f(x) \rangle f(x) f(x)^\top \\
    = & ~ \langle 3f(x)-2b, f(x) \rangle f(x) f(x)^\top
\end{align*}

Maintain $B_{2,2}(x)$ itself:
\begin{align*}
    B_{2,2}(x)
    = & ~ \diag(\langle f(x) - b,f(x) \rangle f(x)) \notag \\
    = & ~ \langle f(x) - b,f(x) \rangle\diag(f(x))
\end{align*}

Merge $B_{1,2}(x)$ and $B_{2,3}(x)$:
\begin{align*}
    B_{1,2}(x) + B_{2,3}(x)
    = & ~ \diag((f(x) - b) \circ f(x)) + \diag(f(x) \circ f(x)) \notag \\
    = & ~ \diag((2f(x) - b) \circ f(x))
\end{align*}

Merge $B_{1,3}(x)$ and $B_{2,4}(x)$:
\begin{align*}
    B_{1,3}(x) + B_{2,4}(x)
    = & ~ (f(x) \circ f(x))f(x)^\top - ((f(x) - b) \circ f(x)) f(x)^\top \notag \\
    = & ~ (f(x) \circ f(x) - f(x) \circ f(x) + b \circ f(x)) f(x)^\top \notag\\
    = & ~ (b \circ f(x)) f(x)^\top
\end{align*}

Merge $B_{1,4}(x)$ and $B_{2,5}(x)$:
\begin{align*}
    B_{1,4}(x) + B_{2,5}(x)
    = & ~ f(x)(f(x) \circ f(x))^\top - f(x)((f(x) - b) \circ f(x))^\top \notag \\
    = & ~ f(x)(f(x)^\top \circ f(x)^\top - f(x)^\top \circ f(x)^\top + b ^\top \circ f(x)^\top) \notag\\
    = & ~ f(x)(b \circ f(x))^\top
\end{align*}

By combining all the above equations, we have
\begin{align*}
    B(x)
    = & ~ \underbrace{ \langle 3f(x)-2b, f(x) \rangle f(x) f(x)^\top }_{B_{1,1} + B_{2,1}} \\
    & ~ + \underbrace{ \langle f(x) - b,f(x) \rangle\diag(f(x)) }_{B_{2,2}} \\
    & ~+ \underbrace{ \diag((2f(x) - b) \circ f(x)) }_{ B_{1,2} + B_{2,3}} \\
    & ~+ \underbrace{ (b \circ f(x)) f(x)^\top }_{B_{1,3} + B_{2,4}} + \underbrace{ f(x)(b \circ f(x))^\top }_{ B_{1,4} + B_{2,5} }
\end{align*}

Thus, we complete the proof.
\end{proof}

%% file: hessian.tex
\section{Hessian is Positive Definite}\label{sec:psd}

In this section, we prove that $\nabla^2 L \succeq 0$ and thus $L$ is convex. 
In Section~\ref{sec:psd:psd_lower_bound}, we find the lower bound of $B(x)$.
To be specific, we split $B(x)$ into several terms and find their lower bounds separately.
In Section~\ref{sec:psd:result}, we use the result of Section~\ref{sec:psd:psd_lower_bound} to prove that lower bound of $\nabla^2 L \succeq 0$ and thus $L$ is convex.

\subsection{PSD Lower Bound
}\label{sec:psd:psd_lower_bound}

For convenient, we define $B(x)$  

\begin{definition}\label{def:B}
We define $B(x)$ as follows
\begin{align*}
    B(x):= & ~ \langle 3 f(x) - 2 b, f(x) \rangle f(x) f(x)^\top \\
    & ~ + (b \circ f(x)) f(x)^\top + f(x) (b \circ f(x))^\top \\
    & ~ + \langle f(x) - b, f(x) \rangle \cdot \diag( f(x) ) \\
    & ~ + \diag( (2f(x) - b) \circ f(x) ) 
\end{align*}
Further, we define
\begin{align*}
    B_{\rank}(x) := & ~ \underbrace{ \langle 3 f(x) - 2 b, f(x) \rangle f(x) f(x)^\top  }_{ :=B_{\rank}^1(x) } + \underbrace{ (b \circ f(x)) f(x)^\top + f(x) (b \circ f(x))^\top }_{ :=B_{\rank}^2(x) } \\
    B_{\diag}(x):= & ~ \underbrace{ \langle f(x) - b, f(x) \rangle \cdot \diag( f(x) ) }_{ := B_{\diag}^1(x)}  + \underbrace{ \diag( (2f(x) - b) \circ f(x) ) }_{ := B_{\diag}^2(x) }
\end{align*}
\end{definition}

\begin{lemma}\label{lem:hessian_L_exp_lower_bound}
If the following conditions hold
\begin{itemize}
    \item $\| f(x) \|_1 = 1$ (see Definition~\ref{def:f}).
    \item Let $B(x) \in \R^{n \times n}$ be defined as Definition~\ref{def:B}.
    \item Let $f(x) \geq {\bf 0}_n$.
    \item Let $b \geq {\bf 0}_n$.
    \item Let $B_{\rank}^1$, $B_{\rank}^2$ be defined as Definition~\ref{def:B}.
    \item Let $B_{\diag}^1$, $B_{\diag}^2$ be defined as Definition~\ref{def:B}.
\end{itemize}
Then we have
\begin{itemize}
\item Part 1.
\begin{align*}
   -0.5 \| b \|_2^2 \cdot f(x) f(x)^\top \preceq B_{\rank}^1(x) \preceq (3 \| f(x) \|_2^2 ) \cdot f(x) f(x)^\top
\end{align*}
\item Part 2. 
\begin{align*}
  -  (1 + \| b \|_{\infty}^2) \cdot f(x) f(x)^\top \preceq B_{\rank}^2(x) \preceq (1+ \| b \|_{\infty}^2) \cdot f(x) f(x)^\top 
\end{align*}
\item Part 3.
\begin{align*}
   -0.25 \| b \|_2^2 \cdot \diag( f(x) ) \preceq B_{\diag}^1 \preceq ( \| f(x) \|_2^2 ) \cdot \diag(f(x))
\end{align*}
\item Part 4.
\begin{align*}
   -0.5 \cdot \diag(b \circ b) \preceq B_{\diag}^2 \preceq 2 \cdot \diag(f(x) \circ f(x))
\end{align*}
\item Part 5. If $\| b \|_1 \leq 1$ and $\| f (x) \|_1 \leq 1$, then we have
\begin{align*}
  -4 I_n \preceq  B(x) \preceq   8 I_n
\end{align*}
\end{itemize}
\end{lemma}
\begin{proof}

Recall that in Definition~\ref{def:B}, we split $B(x)$ into four terms
\begin{align*}
B(x) = B_{\rank}^1 + B_{\rank}^2 + B_{\diag}^1 + B_{\diag}^2,
\end{align*}

where $B_{\rank}^i$ and $B_{\diag}^i$ are defined as
\begin{align*}
B_{\rank}^1 := & ~ \langle 3 f(x) - 2b , f(x) \rangle f(x) f(x)^\top , \\
B_{\rank}^2 := & ~ (b \circ f(x)) f(x)^\top + f(x) (b \circ f(x) )^\top, \\
B_{\diag}^1 := & ~ \langle f(x) - b, f(x) \rangle \diag(f(x)) , \\
B_{\diag}^2 := & ~ \diag( f(x) \circ ( 2f(x) - b) ) .
\end{align*}

{\bf Proof of $B_{\rank}^1$.}

On one hand, we can lower bound the coefficient, we have
\begin{align*}
 \langle 3 f(x) - 2b , f(x) \rangle 
 \geq & ~ 2 \langle f(x) - b, f(x) \rangle \\
 = & ~  2 \langle f(x) - b, f(x) \rangle + 0.5 \| b \|_2^2 - 0.5 \| b \|_2^2 \\
 = & ~ 0.5 \| 2 f(x) - b \|_2^2 - 0.5 \| b \|_2^2 \\
 \geq & ~ - 0.5 \| b \|_2^2 .
\end{align*} 
Thus, 
\begin{align*}
  B_{\rank}^1 \succeq - 0.5 \| b \|_2^2 f(x) f(x)^\top .
\end{align*}
On the other hard, we have
\begin{align*}
 \langle 3 f(x) - 2b , f(x) \rangle 
 = & ~ 3 \| f(x) \|_2^2 - 2 \langle b, f(x) \rangle \\
 \leq & ~  3 \| f(x) \|_2^2 
 \end{align*}
Thus, 
\begin{align*}
    B_{\rank}^1 \preceq 3 \| f(x) \|_2^2  \cdot f(x) f(x)^\top.
\end{align*}

{\bf Proof of $B_{\rank}^2(x)$.}

On one hand, we have
\begin{align*}
B_{\rank}^2 (x)
\succeq & ~ - (b \circ f(x) )^\top (b \circ f(x)) - f(x) f(x)^\top\\ 
= & ~ - (\| b\|_{\infty}^2 +1) \cdot f(x) f(x)^\top,
\end{align*}
where the 1st step follows from Fact~\ref{fac:psd},
, the last step follows from Fact~\ref{fac:psd}.

On the other hand, we have
\begin{align*}
B_{\rank}^2 (x) 
\preceq & ~ (b \circ f(x) )^\top (b \circ f(x)) + f(x) f(x)^\top \\
\preceq & ~ (\| b\|_{\infty}^2 +1) \cdot f(x) f(x)^\top
\end{align*}
where the 1st step follows from
Fact~\ref{fac:psd}
, the 2nd step follows from
Fact~\ref{fac:psd}
.

{\bf Proof of $B_{\diag}^1(x)$.}

For the coefficient, we have
\begin{align*}
\langle f(x) - b, f(x) \rangle
= & ~\langle f(x) - b, f(x) \rangle + \frac{1}{4} \| b \|_2^2   - \frac{1}{4} \| b \|_2^2 \\
= & ~ \| f(x) - \frac{1}{2} b \|_2^2  - \frac{1}{4} \| b \|_2^2 \\
\geq & ~ - \frac{1}{4} \| b \|_2^2
\end{align*}
Thus, we have
\begin{align*}
    B_{\diag}^1  \succeq - \frac{1}{4} \| b \|_2^2 \cdot \diag( f(x) ).
\end{align*}

We can show
\begin{align*}
\langle f(x) - b, f(x) \rangle = & ~ \| f(x) \|_2^2 - \langle b, f(x) \rangle \\
\leq & ~ \| f (x) \|_2^2 
\end{align*}

We have,
\begin{align*}
    B_{\diag}^2 \preceq ( \| f(x) \|_2^2) \cdot \diag(f(x))
\end{align*}

{\bf Proof of $B_{\diag}^2(x)$.}

For the third term, we have
\begin{align*}
B_{\diag}^2 = & ~ \diag( f(x) \circ ( 2f(x) - b) + \frac{1}{2} b \circ b )  - \frac{1}{2} \diag(b \circ b) \\
\succeq & ~ - \frac{1}{2} \diag(b \circ b)  \\
\succeq & ~ - \frac{1}{2} \| b \|_2^2 I_n
\end{align*}
where the 1st step follows from simple algebra, the 2nd step follows from simple algebra, the last step follows from Fact~\ref{fac:psd}.

{\bf Proof of $B(x)$.}
It trivially follows from
\begin{align*}
\| f(x) \|_1 \leq 1, \| b \|_1 \leq 1
\end{align*}
and using Lemma~\ref{lem:basic_f} and Fact~\ref{fac:psd}
\begin{align*}
\max\{ f(x) f(x)^\top , \diag( f(x)), \diag( f(x) \circ f(x)) , \diag (b \circ b) \} \preceq I_n.
\end{align*}

\end{proof}

\subsection{Lower bound on Hessian}\label{sec:psd:result}
The goal of this section is to prove Lemma~\ref{lem:convex}.
\begin{lemma}\label{lem:convex}
If the following conditions hold
\begin{itemize}
    \item Given matrix $A \in \R^{n \times d}$.
    \item Let $L_{\exp}(x)$ be defined as Definition~\ref{def:L_exp}.
    \item Let $L_{\reg}(x)$ be defined as Definition~\ref{def:L_reg}.
    \item Let $L(x) = L_{\exp}(x) + L_{\reg}(x)$. 
    \item Let $W = \diag(w) \in \R^{n \times n}$. Let $W^2 \in \R^{n \times n}$ denote the matrix that $i$-th diagonal entry is $w_{i,i}^2$.
    \item Let $\sigma_{\min}(A)$ denote the minimum singular value of $A$.
    \item Let $l > 0$ denote a scalar.
\end{itemize}
Then, we have
\begin{itemize}
    \item Part 1. If all $i \in [n]$, $w_{i}^2 \geq 4 + l/\sigma_{\min}(A)^2$, then  
    \begin{align*}
    \frac{\d^2 L}{\d x^2} \succeq l \cdot I_d
    \end{align*}
    \item Part 2. If all $i \in [n]$, $w_{i}^2 \geq 100 + l/\sigma_{\min}(A)^2$, then  
    \begin{align*}
       (1-1/10) \cdot ( B(x) + W^2) \preceq  W^2 \preceq (1+1/10) \cdot (B(x) +W^2)
    \end{align*}
\end{itemize}
\end{lemma}

\begin{proof}
    By applying Lemma~\ref{lem:hessian_of_L_x} and Lemma~\ref{lem:B_1_B_2_B}, we have 
    \begin{align*}
        \frac{\d^2 L_{\exp}}{\d x^2}
        = & ~ A^\top B(x) A 
    \end{align*}
    where 
    \begin{align}\label{eq:hessian_total_step_1}
        B(x) \succeq - 4 I_n
    \end{align}
    
    Also, it's trivial that
    \begin{align}\label{eq:hessian_total_step_2}
        \frac{\d^2 L}{\d x^2} = \frac{\d^2 L_{\reg}}{\d x^2} + \frac{\d^2 L_{\exp}}{\d x^2}
    \end{align}

    Thus, by applying Lemma~\ref{lem:L_reg_gradient_hessian}, Eq.~\eqref{eq:hessian_total_step_2} can be written as
    \begin{align*}
        \frac{\d^2 L}{\d x^2} 
        = & ~ A^\top B(x) A + A^\top W^2 A \\
        = & ~ A^\top (B(x) + W^2) A
    \end{align*}

    Let
    \begin{align*}
        D = B(x) + W^2
    \end{align*}
    Then, $\frac{\d^2 L}{\d x^2}$ can be rewrite as
    \begin{align*}
        \frac{\d^2 L}{\d x^2} = A^\top D A
    \end{align*}

    Now, we can bound $D$ as follows
    \begin{align*}
        D
        \succeq & ~ - 4 I_n + w_{\min}^2 I_n \\
        = & ~ ( -4 + w_{\min}^2) I_n \\ 
        \succeq & ~ \frac{l}{\sigma_{\min}(A)^2} I_n
    \end{align*}
    where the 3rd step follows from $w_{\min}^2 \geq 4 + l/\sigma_{\min}(A)^2$, the last step follows from simple algebra.

    Since $D$ is positive definite, then we have 
    \begin{align*}
        A^\top D A \succeq \sigma_{\min}(D) \cdot \sigma_{\min}(A)^2 I_d \succeq l \cdot I_d
    \end{align*}
    Thus, Hessian is positive definite forever and thus the function is convex.
\end{proof}

\section{Hessian is Lipschitz}\label{sec:lipschitz}

In this section, we find the upper bound of $\|\nabla^2 L(x) - \nabla^2 L(y)\|$ and thus proved that $\nabla^2 L$ is lipschitz.
In Section~\ref{sec:lipschitz:core}, we prove that some basic terms satisfy the property of Lipschitz.
In Section~\ref{sec:lipschitz:sum}, we provide a sketch of how we find the bound of $\|\nabla^2 L(x) - \nabla^2 L(y)\|$, to be specific, we split $\|\nabla^2 L(x) - \nabla^2 L(y)\|$ into 8 terms and state that all these terms can be bound by using $\|f(x) - f(y)\|$.
In Section~\ref{sec:lipschitz:step_1}, we use $\|f(x) - f(y)\|$ to bound the first term.
In Section~\ref{sec:lipschitz:step_2}, we use $\|f(x) - f(y)\|$ to bound the second term.
In Section~\ref{sec:lipschitz:step_3}, we use $\|f(x) - f(y)\|$ to bound the third term.
In Section~\ref{sec:lipschitz:step_4}, we use $\|f(x) - f(y)\|$ to bound the fourth term.
In Section~\ref{sec:lipschitz:step_5}, we use $\|f(x) - f(y)\|$ to bound the fifth term.
In Section~\ref{sec:lipschitz:step_6}, we use $\|f(x) - f(y)\|$ to bound the sixth term.
In Section~\ref{sec:lipschitz:step_7}, we use $\|f(x) - f(y)\|$ to bound the seventh term.
In Section~\ref{sec:lipschitz:step_8}, we use $\|f(x) - f(y)\|$ to bound the last term.

\subsection{Main Result}\label{sec:lipschitz:result}
\begin{lemma}\label{lem:lipschitz}
    If the following condition holds
     
    \begin{itemize}
        \item Let $H(x) = \frac{\d^2 L}{\d x^2}$
        \item Let $R > 2$
        \item $\|x \|_2 \leq R, \| y \|_2 \leq R$
        \item $\| A (x-y) \|_{\infty} < 0.01$
        \item $\| A \| \leq R$
        \item $\| b \|_2 \leq R$
        \item $\langle \exp(Ax), {\bf 1}_n \rangle  \geq \beta$ and $\langle \exp(Ay), {\bf 1}_n \rangle \geq \beta$
    \end{itemize}
    Then we have 
    \begin{align*}
        \| H(x) - H(y) \| \leq \beta^{-2} n^{1.5} \exp(20R^2) \cdot \| x- y \|_2
    \end{align*}
\end{lemma}
\begin{proof}
   
    \begin{align*}
        &\|H(x) - H(y)\| \\
        \leq & ~ \| A \| \cdot (2 \| G_1 \| + \| G_2 \| + \cdots + \| G_8 \| ) \| A \| \\
        \leq & ~ R^2 \cdot (2 \| G_1 \| + \| G_2 \| + \cdots + \| G_8 \| ) \\
        \leq & ~ R^2 \cdot 100 R \cdot \| f(x) - f(y) \|_2 \\
        \leq & ~ R^2 \cdot 100 R \cdot \beta^{-2} n^{1.5} \exp(3 R^2) \| x - y \|_2 \\
        \leq & ~ \beta^{-2} n^{1.5} \exp(20 R^2) \| x - y \|_2
    \end{align*}
    where the 1st step follows definition of $G_i$ and matrix spectral norm, the 2nd step follows from $\| A \| \leq R$, the 3rd step follows from Lemma~\ref{lem:sum_G_i}, the 4th step follows from Lemma~\ref{lem:f_lipschitz}, and the last step follows from simple algebra. 
\end{proof}

\subsection{A Core Tool: Lipschitz Property for Several Basic Functions}\label{sec:lipschitz:core}
\begin{lemma}
\label{lem:f_lipschitz}
If the following conditions hold 
\begin{itemize} 
    \item Let $A \in \R^{n \times d}$
    \item Let $b \in \R^n$ satisfy that $\| b \|_1 \leq 1$
    \item Let $\beta \in (0,0.1)$
    \item Let $R \geq 4$
    \item Let $x, y \in \R^d$ satisfy $\| A (x-y) \|_{\infty} < 0.01$
    \item $\| A \| \leq R$
    \item $\langle \exp(Ax) , {\bf 1}_n \rangle \geq \beta$
    \item $\langle \exp(A y), {\bf 1}_n \rangle \geq \beta$
    \item Let $R_f:= \beta^{-2} n^{1.5} \exp(3 R^2)$ 
    \item Let $\alpha(x)$ be defined as Definition~\ref{def:alpha}
    \item Let $c(x)$ be defined as Definition~\ref{def:c}
    \item Let $f(x)$ be defined as Definition~\ref{def:f}
    \item Let $g(x)$ be defined as Definition~\ref{def:g}
\end{itemize}
We have
\begin{itemize}
    \item Part 0. $\| \exp(Ax) \|_2 \leq \sqrt{n} \exp(R^2)$
    \item Part 1. $\| \exp(Ax) - \exp(Ay) \|_2 \leq 2 \sqrt{n} R \exp(R^2) \cdot \| x - y \|_2$
    \item Part 2. $| \alpha(x) - \alpha(y) | \leq \sqrt{n} \cdot \| \exp(Ax) - \exp(Ay) \|_2$
    \item Part 3. $|\alpha(x)^{-1} - \alpha(y)^{-1}| \leq \beta^{-2} \cdot | \alpha(x) - \alpha(y) |$
    \item Part 4. $\| f(x) - f(y) \|_2 \leq R_f \cdot \| x - y \|_2$
    \item Part 5. $\| c(x) - c(y) \|_2 \leq R_f \cdot \| x - y \|_2$
    \item Part 6. $\| g(x) - g(y) \|_2 \leq 16 \cdot R \cdot R_f \cdot \| x - y \|_2$
\end{itemize}
\end{lemma}
\begin{proof}

{\bf Proof of Part 0.}

We can show that
\begin{align*}
    \| \exp(Ax) \|_2
    \leq & ~ \sqrt{n} \cdot \| \exp(Ax) \|_{\infty} \\
    \leq & ~ \sqrt{n} \cdot \exp( \| Ax \|_{\infty} ) \\
    \leq & ~ \sqrt{n} \cdot \exp( \| A x \|_2 ) \\
    \leq & ~ \sqrt{n} \cdot \exp(R^2),
\end{align*}
where the first step follows from Fact~\ref{fac:vector_norm}, the second step follows from Fact~\ref{fac:vector_norm}, the third step follows from Fact~\ref{fac:vector_norm}, and the last step follows from $\|A\| \le R$ and $\|x\|_2 \le R$. 

{\bf Proof of Part 1.}
We have
\begin{align*}
\| \exp(Ax) - \exp(Ay) \|_2 
\leq & ~ \| \exp(Ax) \|_2 \cdot 2 \| A (y-x) \|_{\infty} \notag \\
\leq & ~ \sqrt{n} \exp(R^2) \cdot  2 \| A (y-x) \|_{\infty} \notag \\
\leq & ~ \sqrt{n} \exp(R^2)  \cdot 2 \| A (y-x) \|_2 \notag \\
\leq & ~ \sqrt{n} \exp(R^2)  \cdot 2 \| A \| \cdot \| y - x \|_2 \notag\\
\leq & ~ 2 \sqrt{n} R \exp(R^2) \cdot \|y - x\|_2
\end{align*}
where the 1st step follows  from $\| A (y-x) \|_{\infty} < 0.01$ and Fact~\ref{fac:vector_norm}, 
the 2nd step follows from {\bf Part 0}, 
the 3rd step follows from Fact~\ref{fac:vector_norm}, 
the 4th step follows from Fact~\ref{fac:matrix_norm}, 
the last step follows from $\|A\| \leq R$. 

{\bf Proof of Part 2.}
\begin{align*}
| \alpha(x) - \alpha(y)| = & ~ | \langle \exp(Ax) - \exp(Ay) , {\bf 1}_n \rangle | \\
\leq & ~ \| \exp(Ax) - \exp(Ay) \|_2 \cdot \sqrt{n}
\end{align*}
where the 1st step follows from the definition of $\alpha(x)$,
the 2nd step follows from Cauchy-Schwarz inequality (Fact~\ref{fac:vector_norm}).

{\bf Proof of Part 3.}

We can show that
\begin{align*}
| \alpha(x)^{-1} - \alpha(y)^{-1} | 
= & ~ \alpha(x)^{-1} \alpha(y)^{-1} \cdot | \alpha(x) - \alpha(y) | \\
\leq & ~ \beta^{-2} \cdot | \alpha(x) - \alpha(y) |
\end{align*}
where the 1st step follows from simple algebra, 
the 2nd step follows from $\alpha(x), \alpha(y) \geq \beta$.

{\bf Proof of Part 4.}

We can show that
\begin{align*}
\| f(x) - f(y) \|_2 = & ~ \| \alpha(x)^{-1} \exp(Ax) - \alpha(y)^{-1} \exp(Ay) \|_2 \\
\leq &  ~ \| \alpha(x)^{-1} \exp(Ax) - \alpha(x)^{-1} \exp(A y) \|_2 + \| \alpha(x)^{-1} \exp(Ay) - \alpha(y)^{-1} \exp(Ay) \|_2 \\
\leq & ~ \alpha(x)^{-1} \| \exp(Ax) - \exp(Ay) \|_2 + |\alpha(x)^{-1} - \alpha(y)^{-1}| \cdot \| \exp(Ay) \|_2
\end{align*}
where the 1st step follows from the definition of $f(x)$ and $\alpha(x)$,
the 2nd step follows from triangle inequality,
the 3rd step follows from $\|\alpha A\| \leq |\alpha| \|A\|$(Fact~\ref{fac:matrix_norm}).

For the first term in the above, we have
\begin{align}\label{eq:lipschitz_f_part_1}
 \alpha(x)^{-1} \| \exp(Ax) - \exp(Ay) \|_2
 \leq & ~ \beta^{-1} \| \exp(Ax) - \exp(Ay) \|_2 \notag \\
 \leq & ~ \beta^{-1} \cdot 2 \sqrt{n} R \exp(R^2) \cdot \| x - y \|_2
\end{align}
where the 1st step follows from $\alpha(x) \geq \beta$,
the 2nd step follows from {\bf Part 1}. 

For the second term in the above, we have 
\begin{align}\label{eq:lipschitz_f_part_2}
|\alpha(x)^{-1} - \alpha(y)^{-1}| \cdot \| \exp(Ay) \|_2
\leq & ~ \beta^{-2} \cdot |\alpha(x) - \alpha(y) | \cdot \| \exp(Ay) \|_2 \notag \\
\leq & ~ \beta^{-2} \cdot |\alpha(x) - \alpha(y) | \cdot \sqrt{n} \exp(R^2) \notag \\
\leq & ~ \beta^{-2} \cdot \sqrt{n} \cdot \|\exp(Ax) - \exp(Ay) \|_2 \cdot \sqrt{n} \exp(R^2) \notag \\
\leq & ~ \beta^{-2} \cdot \sqrt{n} \cdot 2 \sqrt{n}  R \exp(R^2) \| x - y \|_2 \cdot \sqrt{n} \exp(R^2) \notag \\
= & ~ \beta^{-2} \cdot 2 n^{1.5} R \exp(2R^2) \| x - y \|_2
\end{align}
where the 1st step follows from the result of {\bf Part 3},
the 2nd step follows from {\bf Part 0},
the 3rd step follows from the result of {\bf Part 2},
the 4th step follows from {\bf Part 1}, and the last step follows from simple algebra.

Combining Eq.~\eqref{eq:lipschitz_f_part_1} and Eq.~\eqref{eq:lipschitz_f_part_2} together, we have
\begin{align*}
\| f(x)  - f(y) \|_2 \leq & ~ \beta^{-1} \cdot 2 \sqrt{n} R \exp(R^2) \cdot \| x - y \|_2 + \beta^{-2} 2 n^{1.5} R \exp(2R^2) \| x - y \|_2 \\
\leq & ~ 3 \beta^{-2} n^{1.5} R \exp(2R^2) \| x - y \|_2 \\
\leq & ~ \beta^{-2} n^{1.5} \exp(3 R^2) \| x - y \|_2
\end{align*}
where the 1st step follows from the bound of the first term and the second term,
the 2nd step follows from $\beta^{-1} \geq 1$ and $n > 1$ trivially,
the 3rd step follows from simple algebra.

{\bf Proof of Part 5.}
We have
\begin{align*}
\| c(x) - c(y) \|_2 = \| f(x) - f(y) \|_2 \leq R_f \cdot \| x - y \|_2,
\end{align*}
the first step follows from the definition of $c(x)$, the last step follows from {\bf Part~4} and definition of $R_f$. 
{\bf Proof of Part 6.}

Using Lemma~\ref{lem:gradient_lipschitz}, we have
\begin{align*}
\| g(x) - g(y) \|_2 \leq & ~ 8 \| A \| R_f \cdot 2 \| x - y \|_2 \\
\leq & ~ 18 R R_f \cdot \| x- y \|_2,
\end{align*}
where the second step follows from $\|A\| \le R$. 

Thus, we complete the proof.
\end{proof}

\subsection{Summary of 
Eight Steps}\label{sec:lipschitz:sum}
\begin{lemma}\label{lem:sum_G_i}
If the following conditions hold
\begin{itemize}
    \item $G_1 = \| f(x) \|_2^2 f(x) f(x)^\top - \| f(y) \|_2^2 f(y) f(y)^\top$
    \item $G_2 = \langle f(x), b\rangle f(x) f(x)^\top - \langle f(y), b \rangle f(y) f(y)^\top$
    \item $G_3 = \langle f(x), f(x) \rangle \diag(f(x)) - \langle f(y) , f(y) \rangle \diag( f(y) )$
    \item $G_4 = \langle f(x), b \rangle \diag( f(x)) - \langle f(y), b \rangle \diag(f(y))$
    \item $G_5 = \diag(f(x) \circ (f(x) - b)) - \diag( f(y) \circ (f(y) - b) )$
    \item $G_6 = \diag(f(x) \circ f(x)) - \diag( f(y) \circ f(y) )$
    \item $G_7 = f(x) (f(x) \circ b)^\top - f(y) (f(y) \circ b)^\top$ 
    \item $G_8 = (f(x) \circ b) f(x)^\top - ( f(y) \circ b) f(y)^\top$
\end{itemize}
We have
\begin{align*}
    \| G_1 \| + \sum_{i=1}^8 \| G_i \| \leq 100 R \cdot \| f(x) - f(y) \|_2
\end{align*}
\end{lemma}
\begin{proof}
The proof directly follows from applying Lemma~\ref{lem:G_1}, Lemma~\ref{lem:G_2}, Lemma~\ref{lem:G_3}, Lemma~\ref{lem:G_4}, Lemma~\ref{lem:G_5}, Lemma~\ref{lem:G_6}, Lemma~\ref{lem:G_7}, Lemma~\ref{lem:G_8}.
\end{proof}

\subsection{Lipschitz Calculations: Step 1. Lipschitz for Matrix Function \texorpdfstring{$\| f(x) \|_2^2 f(x) f(x)^\top$}{}}\label{sec:lipschitz:step_1}

\begin{lemma}\label{lem:G_1}
If the following condition holds
\begin{itemize}
    \item $G_1 = \| f(x) \|_2^2 f(x) f(x)^\top - \| f(y) \|_2^2 f(y) f(y)^\top $
\end{itemize}
Then 
\begin{align*}
    \| G_1 \| \leq 4 \| f(x) - f(y) \|_2
\end{align*}
\end{lemma}
\begin{proof}
We define
\begin{align*}
G_{1,1}:= & ~ \langle f(x), f(x) \rangle  f(x) f(x)^\top -  \langle f(x) , f(y ) \rangle f(x) f(x)^\top \\
G_{1,2}:= & ~ \langle f(x) , f(y ) \rangle f(x) f(x)^\top  - \langle f(y ) , f(y) \rangle f(x) f(x)^\top \\
G_{1,3}:= & ~ \langle f(y ) , f(y) \rangle f(x) f(x)^\top - \langle f(y ) , f(y) \rangle f(y) f(x)^\top \\
G_{1,4}:= & ~ \langle f(y ) , f(y) \rangle f(y) f(x)^\top -  \langle f(y ) , f(y) \rangle f(y) f(y)^\top
\end{align*}
We have
\begin{align*}
G_1
= G_{1,1} + G_{1,2} + G_{1,3} + G_{1,4}
\end{align*}
Let us only prove for $G_{1,1}$, the others are similar,
\begin{align*}
\| G_{1,1} \| 
\leq & ~ | \langle f(x), f(x) - f(y) \rangle | \cdot \| f(x) f(x)^\top \| \\
\leq & ~ \| f(x) \|_2 \cdot \| f(x)  - f(y ) \|_2 \cdot \| f(x) f(x)^\top  \| \\
= & ~ \| f(x) \|_2 \cdot \| f(x)  - f(y ) \|_2 \cdot \| f(x) \|_2^2 \\
\leq & ~ \| f(x) - f(y) \|_2
\end{align*}
where the 1st step follows from Fact~\ref{fac:matrix_norm}, the 2nd step follows from $|\langle a,b\rangle|\leq \| a \|_2 \| b\|_2$ (Fact~\ref{fac:vector_norm}), 
the 3rd step follows from $aa^\top \preceq \|a\|_2^2 I_n$(Fact~\ref{fac:vector_norm}),
the last step follows from $\| f(x) \|_2 \leq \| f(x) \|_1 \leq 1$ (Lemma~\ref{lem:basic_f}).

It is obvious that for each $i \in [4]$, we have
\begin{align*}
\| G_{1,i} \| \leq & ~ \| f(x) - f(y) \|_2  \max\{ \| f(x) \|_2 , \| f(y)\|_2 \}^3  \\
\leq & ~ \| f(x) - f(y) \|_2 
\end{align*}
where the last step follows from $\| f(x) \|_2 \leq \| f(x) \|_1 \leq 1$.
\end{proof}

\subsection{Lipschitz Calculations: Step 2. Lipschitz for Matrix Function \texorpdfstring{$\langle f(x), b \rangle f(x) f(x)^\top$}{}}\label{sec:lipschitz:step_2}

\begin{lemma}\label{lem:G_2}
If the following condition holds
\begin{itemize}
    \item $G_2:= \langle f(x), b \rangle f(x) f(x)^\top - \langle f(y), b \rangle f(y) f(y)^\top$
\end{itemize}
Then we have
\begin{align*}
    \| G_2 \| \leq 3 \| f(x) - f(y) \|_2 \cdot \| b \|_2
\end{align*}
\end{lemma}
\begin{proof}
 
We define
\begin{align*}
    G_{2,1} := & ~ \langle f(x), b \rangle f(x) f(x)^\top - \langle f(x), b \rangle f(y) f(x)^\top \\
    G_{2,2} := & ~ \langle f(x), b \rangle f(y) f(x)^\top - \langle f(x), b \rangle f(y) f(y)^\top\\
    G_{2,3} := & ~ \langle f(x), b \rangle f(y) f(y)^\top  - \langle f(y), b \rangle f(y) f(y)^\top
\end{align*}
Then it's apparent that
\begin{align*}
    G_2 = G_{2,1} + G_{2,2} + G_{2,3}
\end{align*}
Since $G_{2,1},G_{2,2},G_{2,3}$ are similar,  
we only have to bound $\|G_{2,1}\|$:
\begin{align*}
    \|G_{2,1}\| = & ~ \|\langle f(x), b \rangle f(x) f(x)^\top - \langle f(x), b \rangle f(y) f(x)^\top\| \\
    = & ~ \|\langle f(x), b \rangle (f(x) - f(y)) f(x)^\top\| \\
    \leq & ~ |\langle f(x), b \rangle | \cdot \| (f(x) - f(y) ) f(x)^\top \| \\
    \leq & ~ |\langle f(x), b \rangle | \cdot \|f(x) - f(y) \|_2 \cdot \| f(x) \|_2 \\
    \leq & ~ \|f(x)\|_2^2 \cdot \|b\|_2 \|f(x) - f(y)\|_2 \\
    \leq & ~ \|f(x) - f(y)\|_2 \cdot \|b\|_2
\end{align*}
where the 1st step follows from the definition of $G_{2,1}$,
the 2nd step follows from simple algebra,
the 3rd step follows from Fact~\ref{fac:matrix_norm},
the 4th step follows from $\|ab^\top\| \leq \|a\|_2 \|b\|_2$(Fact~\ref{fac:matrix_norm}),
the 5th step follows from $\langle a,b \rangle \leq \|a\|_2\|b\|_2$(Fact~\ref{fac:vector_norm}),
the last step follows from $\|f(x)\|_2 \leq \|f(x)\|_1 \leq 1$.

Thus, we have
\begin{align*}
    \|G_2\| \leq 3 \|f(x) - f(y)\| \|b\|_2
\end{align*}
\end{proof}

\subsection{Lipschitz Calculations: Step 3. Lipschitz for Matrix Function \texorpdfstring{$f(x) f(x)^\top \diag(f(x))$}{}}\label{sec:lipschitz:step_3}
\begin{lemma}\label{lem:G_3}
If the following condition holds
\begin{itemize}
    \item $G_3:= \langle f(x), f(x) \rangle \diag(f(x)) - \langle f(y), f(y) \rangle \diag(f(y))$
\end{itemize}
Then we have
\begin{align*}
    \| G_3 \| \leq 3 \| f(x) - f(y) \|_2
\end{align*}
\end{lemma}
\begin{proof}
We define
\begin{align*}
    G_{3,1} : & = \langle f(x), f(x) \rangle \diag(f(x)) - \langle f(x), f(y) \rangle \diag(f(x)) \\
    G_{3,2} : & = \langle f(x), f(y) \rangle \diag(f(x)) - \langle f(x), f(y) \rangle \diag(f(y)) \\
    G_{3,3} : & = \langle f(x), f(y) \rangle \diag(f(y)) - \langle f(y), f(y) \rangle \diag(f(y))
\end{align*}
Thus, it's trivial that
\begin{align*}
    G_3 = G_{3,1} + G_{3,2} + G_{3,3}
\end{align*}
Since $G_{3,1},G_{3,2},G_{3,3}$ are similar, we only need to bound $\|G_{3,1}\|$:
\begin{align*}
    \|G_{3,1}\|
    = & ~ \|\langle f(x), f(x) \rangle \diag(f(x)) - \langle f(x), f(y) \rangle \diag(f(x))\| \\
    = & ~ \|\langle f(x),f(x) - f(y) \rangle \diag(f(x))\| \\
    \leq & ~ \|f(x)^\top\|_2 \|f(x) - f(y)\|_2 \|\diag(f(x))\| \\
    = & ~ \|f(x)\|_2^2 \|f(x) - f(y)\|_2 \\
    \leq & ~ \|f(x) - f(y)\|_2
\end{align*}
where the 1st step follows from the definition of $G_{3,1}$,
the 2nd step follows from simple algebra,
the 3rd step follows from $\|\alpha A\| \leq |\alpha|\|A\|$(Fact~\ref{fac:matrix_norm}), $\langle a,b \rangle \leq \|a\|_2 \|b\|_2$(Fact~\ref{fac:vector_norm}), and $\|ab\| \leq \|a\|\|b\|$(Fact~\ref{fac:matrix_norm}),
the 4th step follows from $\|\diag(f(x))\| = \|f(x)\|_2$,
the last step follows from $\|f(x)\|_2 \leq \|f(x)\|_1 \leq 1$ (Fact~\ref{fac:vector_norm}).

Thus, we have
\begin{align*}
    \|G_8\|
    = & ~ \|G_{8,1} + G_{8,2} + G_{3,3}\| \\
    \leq & ~ \|G_{8,1}\| + \|G_{8,2}\| + \|G_{3,3}\| \\
    = & ~ 3 \| f(x) - f(y) \|_2
\end{align*}
where the 1st step follows from the definition of $G_3$,
the 2nd step follows from Fact~\ref{fac:matrix_norm},
the last step follows from the bound of $\|G_{3,1}\|$,$\|G_{3,2}\|$ and $\|G_{3,3}\|$.
\end{proof}

\subsection{Lipschitz Calculations: Step 4. Lipschitz for Matrix Function \texorpdfstring{$\langle f(x), b \rangle \diag(f(x))$}{}}\label{sec:lipschitz:step_4}
\begin{lemma}\label{lem:G_4}
If the following condition holds
\begin{itemize}
    \item $G_4:= \langle f(x), b \rangle \diag(f(x)) - \langle f(y), b \rangle \diag( f(y) )$
\end{itemize}
Then we have
\begin{align*}
    \| G_4 \| \leq 2 \| f(x) - f(y) \|_2 \| b \|_2
\end{align*}
\end{lemma}
\begin{proof}
We define:
\begin{align*}
    G_{4,1} : & = \langle f(x), b \rangle \diag(f(x)) - \langle f(y), b \rangle \diag( f(x) ) \\
    G_{4,2} : & = \langle f(y), b \rangle \diag(f(x)) - \langle f(y), b \rangle \diag( f(y) )
\end{align*}

Thus, it's trivial that
\begin{align*}
    G_4 = G_{4,1} + G_{4,2}
\end{align*}

Since $G_{4,1}$ and $G_{4,2}$ are similar, we only need to bound $\|G_{4,1}\|$:
\begin{align*}
    \|G_{4,1}\|
    = & ~ \|\langle f(x), b \rangle \diag(f(x)) - \langle f(y), b \rangle \diag( f(x) ) \| \\
    = & ~ \|b^\top (f(x) - f(y)) \diag(f(x))\| \\
    \leq & ~ \|b^\top\|_2\|f(x) - f(y)\|_2\|\diag(f(x)\| \\
    \leq & ~ \|b\|_2 \|f(x) - f(y)\|_2 \|f(x)\|_2 \\
    \leq & ~ \|f(x) - f(y)\|_2\|b\|_2
\end{align*}
where the 1st step follows from the definition of $G_{4,1}$,
the 2nd step follows from simple algebra,
the 3rd step follows from $\|ab\| \leq \|a\|\|b\|$(Fact~\ref{fac:matrix_norm}) and ,
the 4th step follows from $\|\diag(x)\| \leq \|x\|_{\infty} \leq \|x\|_2$(Fact~\ref{fac:vector_norm}),
the last step follows from $\|f(x)\|_2 \leq \|f(x)\|_1 \leq 1$(Fact~\ref{fac:vector_norm}).

Thus, we have
\begin{align*}
    \|G_4\|
    = & ~ \|G_{4,1} + G_{4,2}\| \\
    \leq & ~ \|G_{4,1}\| + \|G_{4,2}\| \\
    = & ~ 2 \| f(x) - f(y) \|_2 \cdot \|b \|_2
\end{align*}
where the 1st step follows from the definition of $G_4$,
the 2nd step follows from Fact~\ref{fac:matrix_norm},
the last step follows from the bound of $\|G_{4,1}\|$ and $\|G_{4,2}\|$.
\end{proof}

\subsection{Lipschitz Calculations: Step 5. Lipschitz for Matrix Function \texorpdfstring{$\diag( f(x) \circ (f(x) - b) )$}{}}\label{sec:lipschitz:step_5}

\begin{lemma}\label{lem:G_5}
If the following condition holds
\begin{itemize}
    \item $ G_5:= \diag( f(x) \circ (f(x) - b) ) - \diag( f(y) \circ (f(y) - b) )$
\end{itemize}
Then we have
\begin{align*}
    \| G_5 \| \leq 2 \| f(x) - f(y) \|_2 + \| f(x) - f(y) \| \cdot \| b \|_2
\end{align*}
 \end{lemma}
 \begin{proof}
We define:
\begin{align*}
    G_{5,1} : & = \diag( f(x) \circ (f(x) - b) ) - \diag( f(x) \circ (f(y) - b) ) \\
    G_{5,2} : & = \diag( f(x) \circ (f(y) - b) ) - \diag( f(y) \circ (f(y) - b) )
\end{align*}

Then, it's trivial that
\begin{align*}
    G_5 = G_{5,1} + G_{5,2}
\end{align*}

Bound $\|G_{5,1}\|$:
\begin{align*}
    \|G_{5,1}\|
    = & ~ \|\diag( f(x) \circ (f(x) - b) ) - \diag( f(x) \circ (f(y) - b) )\| \\
    = & ~ \|\diag(f(x))\diag(f(x) - f(y))\| \\
    \leq & ~ \|\diag(f(x))\|\|\diag(f(x) - f(y))\| \\
    \leq & ~ \|f(x)\|_2 \|f(x) - f(y)\|_2 \\
    \leq & ~ \|f(x) - f(y)\|_2
\end{align*}
where the 1st step follows from the definition of $G_{5,1}$,
the 2nd step follows from Fact~\ref{fac:circ_diag},
the 3rd step follows from $\|ab\| \leq \|a\|\|b\|$(Fact~\ref{fac:matrix_norm}),
the 4th step follows from $\|\diag(a)\| \leq \|a\|_{\infty} \leq \|a\|_2$(Fact~\ref{fac:vector_norm}),
the last step follows from $\|f(x)\|_2 \leq \|f(x)\|_1 \leq 1$.

Bound $\|G_{5,2}\|$:
\begin{align*}
    \|G_{5,2}\|
    = & ~ \|\diag( f(x) \circ (f(y) - b) ) - \diag( f(y) \circ (f(y) - b) )\| \\
    = & ~ \|\diag(f(x) - f(y)) \diag(f(y) - b)\| \\
    \leq & ~ \|\diag(f(x) - f(y))\|\|\diag(f(y) - b)\| \\
    \leq & ~ \|f(x) - f(y)\|_2 \|f(y)\|_2 + \|f(x) - f(y)\|_2 \|b\|_2 \\
    \leq & ~ \|f(x) - f(y)\|_2 + \|f(x) - f(y)\|_2 \|b\|_2
\end{align*}
where the 1st step follows from the definition of $G_{5,2}$,
the 2nd step follows from Fact~\ref{fac:circ_diag},
the 3rd step follows from Fact~\ref{fac:matrix_norm},
the 4th step follows from $\|\diag(a)\| \leq \|a\|_{\infty} \leq \|a\|_2$(Fact~\ref{fac:vector_norm}),
the last step follows from $\|f(x)\|_2 \leq \|f(x)\|_1 \leq 1$.

Thus, we have
\begin{align*}
    \|G_5\|
    = & ~ \|G_{5,1} + G_{5,2}\| \\
    \leq & ~ \|G_{5,1}\| + \|G_{5,2}\| \\
    \leq & ~ 2 \| f(x) - f(y) \|_2 + \| f(x) - f(y) \|_2 \cdot \| b \|_2
\end{align*}
where the 1st step follows from the definition of $G_5$,
the 2nd step follows from Fact~\ref{fac:circ_diag},
the 3rd step follows from the bound of $\|G_{5,1}\|$ and $\|G_{5,2}\|$.
\end{proof}

\subsection{Lipschitz Calculations: Step 6. Lipschitz for Matrix Function \texorpdfstring{$\diag(f(x) \circ f(x))$}{}}\label{sec:lipschitz:step_6}

\begin{lemma}\label{lem:G_6}
If the following condition holds 
\begin{itemize}
    \item $ G_6:= \diag(f(x) \circ f(x)) - \diag( f(y) \circ f(y) ) $
\end{itemize}
Then we have
\begin{align*}
\| G_6 \| \leq 2 \| f(x) - f(y) \|_2
\end{align*}
 \end{lemma}
 \begin{proof}
We define:
\begin{align*}
    G_{6,1} : & = \diag(f(x) \circ f(x)) - \diag( f(x) \circ f(y) ) \\
    G_{6,2} : & = \diag(f(x) \circ f(y)) - \diag( f(y) \circ f(y) )
\end{align*}

Then, it's trivial that
\begin{align*}
    G_6 = G_{6,1} + G_{6,2}
\end{align*}

Since, $G_{6,1}$ and $G_{6,2}$ are similar, we only need to bound $\|G_{6,1}\|$:
\begin{align*}
    \|G_{6,1}\|
    = & ~ \|\diag(f(x) \circ f(x)) - \diag( f(x) \circ f(y) )\| \\
    = & ~ \|\diag(f(x))(\diag(f(x) - \diag(f(y)))\| \\
    \leq & ~ \|f(x)\|_2 \|f(x) - f(y)\|_2 \\
    \leq & ~ \|f(x) - f(y)\|
\end{align*}
where the 1st step follows from the definition of $G_{6,1}$,
the 2nd step follows from Fact~\ref{fac:circ_diag},
the 3rd step follows from $\|ab\| \leq \|a\|\|b\|$(Fact~\ref{fac:matrix_norm}) and $\|\diag(a)\| \leq \|a\|_{\infty} \leq \|a\|_2$(Fact~\ref{fac:vector_norm}),
the last step follows from $\|f(x)\|_2 \leq \|f(x)\|_1 \leq 1$.

Thus, we have
\begin{align*}
    \|G_6\|
    = & ~ \|G_{6,1} + G_{6,2}\| \\
    \leq & ~ \|G_{6,1}\| + \|G_{6,2}\| \\
    = & ~ 2 \| f(x) - f(y) \|_2
\end{align*}
where the 1st step follows from the definition of $G_6$,
the 2nd step follows from Fact~\ref{fac:matrix_norm},
the last step follows from the bound of $\|G_{6,1}\|$ and $\|G_{6,2}\|$.
\end{proof}

 \subsection{Lipschitz Calculations: Step 7. Lipschitz for Matrix Function \texorpdfstring{$f(x) (f(x) \circ b)^\top$}{}}\label{sec:lipschitz:step_7}

\begin{lemma}\label{lem:G_7}
If the following condition holds
\begin{itemize}
    \item $ G_7:= f(x) (f(x) \circ b )^\top - f(y) (f(y) \circ b)^\top$
\end{itemize}
Then, we have
\begin{align*}
\| G_7 \| \leq 2 \| f(x) - f(y) \|_2 \cdot \| b \|_2
\end{align*}
 \end{lemma}
 \begin{proof}
We define:
\begin{align*}
    G_{7,1} : & = f(x) (f(x) \circ b )^\top - f(x) (f(y) \circ b)^\top \\
    G_{7,2} : & = f(x) (f(y) \circ b )^\top - f(y) (f(y) \circ b)^\top
\end{align*}

Since $G_{7,1}$ and $G_{7,2}$ are similar, we only need to bound $\|G_{7,1}\|$:
\begin{align*}
    \|G_{7,1}\|
    = & ~ \|f(x) (f(x) \circ b )^\top - f(x) (f(y) \circ b)^\top\| \\
    = & ~ \|f(x)((f(x) - f(y)) \circ b)^\top\| \\
    \leq & ~ \|f(x)\|_2 \|(f(x) - f(y)) \circ b\|_2 \\
    \leq & ~ \|f(x) - f(y)\|_2 \|b\|_2
\end{align*}
where the 1st step follows from the definition of $G_{7,1}$,
the 2nd step follows from simple algebra,
the 3rd step follows from $\|ab^\top\| \leq \|a\|_2 \|b\|_2$(Fact~\ref{fac:matrix_norm}) and $\|a^\top\|_2 = \|a\|_2$,
the last step follows from $\|a \circ b\|_2 \leq \|a\|_{\infty} \|b\| \leq \|a\|_2\|b\|_2$(Fact~\ref{fac:vector_norm}) and $\|f(x)\|_2 \leq \|f(x)\|_1 \leq 1$.

Thus, we have
\begin{align*}
    \|G_7\|
    = & ~ \|G_{7,1} + G_{7,2}\| \\
    \leq & ~ \|G_{7,1}\| + \|G_{7,2}\| \\
    = & ~ 2 \| f(x) - f(y) \|_2 \cdot \|b \|_2
\end{align*}
where the 1st step follows from the definition of $G_7$,
the 2nd step follows from Fact~\ref{fac:matrix_norm},
the last step follows from the bound of $\|G_{7,1}\|$ and $\|G_{7,2}\|$.
\end{proof}

 \subsection{Lipschitz Calculations: Step 8. Lipschitz for Matrix Function \texorpdfstring{$(f(x)\circ b) f(x)^\top$}{}}\label{sec:lipschitz:step_8}

\begin{lemma}\label{lem:G_8}
If the following condition holds
\begin{itemize}
    \item $ G_8 :=  (f(x) \circ b) f(x)^\top - ( f(y) \circ b) f(y)^\top $
\end{itemize}
Then we have
\begin{align*}
\| G_8 \| \leq 2 \| f(x) - f(y) \|_2 \cdot \|b \|_2
\end{align*}
 \end{lemma}
 \begin{proof}
We define:
\begin{align*}
    G_{8,1} : & = (f(x) \circ b) f(x)^\top - ( f(x) \circ b) f(y)^\top \\
    G_{8,2} : & = (f(x) \circ b) f(y)^\top - ( f(y) \circ b) f(y)^\top
\end{align*}

Then, it's trivial that
\begin{align*}
    G_8 = G_{8,1} + G_{8,2}
\end{align*}

Since $G_{8,1}$ and $G_{8,2}$ are similar, we only need to bound $\|G_{8,1}\|$:
\begin{align*}
    \|G_{8,1}\|
    = & ~ \|(f(x) \circ b) f(x)^\top - ( f(x) \circ b) f(y)^\top\| \\
    = & ~ \|(f(x) \circ b)(f(x) - f(y))^\top\| \\
    \leq & ~ \|f(x) \circ b\|_2 \|f(x) - f(y)\|_2 \\
    \leq & ~ \|f(x)\|_2 \|b\|_2 \|f(x) - f(y)\|_2 \\
    \leq & ~ \|f(x) - f(y)\|_2 \|b\|_2
\end{align*}
where the 1st step follows from the definition of $G_{8,1}$,
the 2nd step follows from simple algebra,
the 3rd step follows from $\|ab^\top\| \leq \|a\|_2\|b\|_2$(Fact~\ref{fac:matrix_norm}),
the 4th step follows from $\|a \circ b\|_2 \leq \|a\|_{\infty} \|b\| \leq \|a\|_2\|b\|_2$(Fact~\ref{fac:vector_norm}),
the last step follows from $\|f(x)\|_2 \leq \|f(x)\|_1 \leq 1$ (Lemma~\ref{lem:basic_f}).

Thus, we have
\begin{align*}
    \|G_8\|
    = & ~ \|G_{8,1} + G_{8,2}\| \\
    \leq & ~ \|G_{8,1}\| + \|G_{8,2}\| \\
    = & ~ 2 \| f(x) - f(y) \|_2 \cdot \|b \|_2
\end{align*}
where the 1st step follows from the definition of $G_8$,
the 2nd step follows from Fact~\ref{fac:matrix_norm},
the last step follows from the bound of $\|G_{8,1}\|$ and $\|G_{8,2}\|$.
\end{proof}

%% file: result.tex
\section{Approximate Newton Method}\label{sec:newton}

In this section, we provide an approximate version of the newton method for convex optimization.
In Section~\ref{sec:newton:definitions}, we state some assumptions of the traditional newton method and the exact update rule of the traditional algorithm.
In Section~\ref{sec:newton:approximation}, we provide the approximate update rule of the approximate newton method, we also implement a tool for compute the approximation of $\nabla^2 L$ and use some lemmas from \cite{lsz23} to analyze the approximate newton method. In Section~\ref{sec:newton:beta}, we prove a lower bound on $\beta$. In Section~\ref{sec:newton:M}, we prove an upper bound on $M$.

\subsection{Definition and Update Rule}\label{sec:newton:definitions}
Here in this section, we focus on the local convergence of the Newton method. We consider the following target function
\begin{align*}
    \min_{x \in \R^d } L(x)
\end{align*}
with these assumptions:
\begin{definition}[$(l,M)$-good Loss function]\label{def:f_ass}
For a function $L : \R^d \rightarrow \R$, we say $L$ is $(l,M)$-good it satisfies the following conditions,
\begin{itemize}
    \item {\bf $l$-local Minimum.}  
    We define $l >0$ to be a positive scalar. If there exists a vector $x^* \in \R^d$ such that the following holds
    \begin{itemize}
        \item $\nabla L(x^*) = {\bf 0}_d$.
        \item $\nabla^2 L(x^*) \succeq l \cdot I_d$.
    \end{itemize}
    \item {\bf Hessian is $M$-Lipschitz.} If there exists a positive scalar $M>0$ such that
    \begin{align*}
        \| \nabla^2 L(y) - \nabla^2 L(x) \| \leq M \cdot \| y - x \|_2 
    \end{align*}
    \item {\bf Good Initialization Point.} Let $x_0$ denote the initialization point. If $r_0:=\| x_0 -x_*\|_2$ satisfies
    \begin{align*}
        r_0 M \leq 0.1 l
    \end{align*}    
\end{itemize}
\end{definition}

We define gradient and Hessian as follows
\begin{definition}[Gradient and Hessian] 
The gradient $g : \R^d \rightarrow \R^d$ of the loss function is defined as 
\begin{align*}
    g(x) := \nabla L(x)
\end{align*}
The Hessian $H : \R^d \rightarrow \R^{d \times d}$ of the loss function is defined as,
\begin{align*}
    H(x) := \nabla^2 L(x)
\end{align*}
\end{definition}

With the gradient function $g: \R^d \rightarrow \R^d$ and the Hessian matrix $H : \R^d \rightarrow \R^{d \times d}$, we define the exact process of the Newton method as follows: 
\begin{definition}[Exact update of the Newton method]\label{def:exact_update_variant}
\begin{align*}
    x_{t+1} = x_t - H(x_t)^{-1} \cdot g(x_t)
\end{align*}
\end{definition}

\subsection{Approximate of Hessian and Update Rule}\label{sec:newton:approximation}
In many real-world tasks, it is very hard and expensive to compute exact $\nabla^2 L(x_t )$ or $(\nabla^2 L(x_t))^{-1}$. Thus, it is natural to consider the approximated computation of the gradient and Hessian. The computation is defined as

\begin{definition}[Approximate Hessian]\label{def:wt_H}
For any Hessian $H(x_t) \in \R^{d \times d}$, we define the approximated Hessian $\wt{H}(x_t) \in \R^{d \times d}$ to be a matrix such that the following holds,
\begin{align*}
 (1-\epsilon_0) \cdot H(x_t) \preceq \wt{H}(x_t) \preceq (1+\epsilon_0) \cdot H(x_t) .
\end{align*}
\end{definition}

In order to get the approximated Hessian $\wt{H}(x_t)$ efficiently, here we state a standard tool (see Lemma~4.5 in \cite{dsw22}).
\begin{lemma}[\cite{dsw22,syyz22}]\label{lem:subsample}
Let $\epsilon_0 = 0.01$ be a constant precision parameter. 
Let $A \in \R^{n \times d}$ be a real matrix, then for any positive diagonal (PD) matrix $D \in \R^{n \times n}$, there exists an algorithm which runs in time
\begin{align*}
O( (\nnz(A) + d^{\omega} ) \poly(\log(n/\delta)) )
\end{align*}
and it outputs an $O(d \log(n/\delta))$ sparse diagonal matrix $\wt{D} \in \R^{n \times n}$ for which 
\begin{align*}
(1- \epsilon_0) A^\top D A \preceq A^\top \wt{D} A \preceq (1+\epsilon_0) A^\top D A.
\end{align*}
Note that, $\omega$ denotes the exponent of matrix multiplication, currently $\omega \approx 2.373$ \cite{w12,lg14,aw21}.
\end{lemma}

Following the standard of Approximate Newton Hessian literature \cite{a00,jkl+20,bpsw21,szz21,hjs+22,lsz23}, we consider the following.
\begin{definition}[Approximate update]\label{def:update_x_k+1}

We consider the following process
\begin{align*}
    x_{t+1} = x_t  - \wt{H}(x_t)^{-1} \cdot  g(x_t)  .
\end{align*}
\end{definition}

We state a tool from prior work,
\begin{lemma}[Iterative shrinking Lemma, Lemma 6.9 on page 32 of \cite{lsz23}]\label{lem:one_step_shrinking}
If the following condition hold
\begin{itemize}
    \item Loss Function $L$ is $(l,M)$-good (see  Definition~\ref{def:f_ass}). 
    \item Let $\epsilon_0 \in (0,0.1)$ (see Definition~\ref{def:wt_H}). 
    \item Let $r_t:= \| x_t - x^* \|_2$.
    \item Let $\ov{r}_t: = M \cdot r_t$
\end{itemize}
 Then we have  
\begin{align*}
r_{t+1} \leq 2 \cdot (\epsilon_0 + \ov{r}_t/( l - \ov{r}_t ) ) \cdot r_t.
\end{align*} 
\end{lemma}

Let $T$ denote the total number of iterations of the algorithm, to apply Lemma~\ref{lem:one_step_shrinking}, we will need the following induction hypothesis lemma. This is very standard in the literature, see \cite{lsz23}.
\begin{lemma}[Induction hypothesis, Lemma 6.10 on page 34 of \cite{lsz23}]\label{lem:newton_induction}
For each $i \in [t]$, we define $r_i:= \| x_i - x^* \|_2$. 
If the following condition hold
\begin{itemize}
    \item $\epsilon_0 = 0.01$ (see Definition~\ref{def:wt_H} for $\epsilon_0$)
    \item $r_{i} \leq 0.4 \cdot r_{i-1}$, for all $i \in [t]$
    \item $M \cdot r_i \leq 0.1 l$, for all $i \in [t]$ (see Definition~\ref{def:f_ass} for $M$)
\end{itemize}
Then we have
\begin{itemize}
    \item $r_{t+1} \leq 0.4 r_t$
    \item $M \cdot r_{t+1} \leq 0.1 l$
\end{itemize}
\end{lemma}

\subsection{Lower bound on \texorpdfstring{$\beta$}{}}\label{sec:newton:beta}

\begin{lemma}\label{lem:beta}
If the following conditions holds
\begin{itemize}
    \item $\| A \| \leq R$
    \item $\| x \|_2 \leq R$
    \item Let $\beta$ be lower bound on $\langle \exp(Ax), {\bf 1}_n \rangle$
\end{itemize}
Then we have 
\begin{align*}
    \beta \geq \exp(-R^2)
\end{align*}
\end{lemma}
\begin{proof}
We have
\begin{align*}
\langle \exp(Ax) ,{\bf 1}_n \rangle 
\geq & ~ \max_{i \in [n]} \exp( - | (Ax)_i| ) \\
\geq & ~ \exp(-\| A x \|_{\infty}) \\
\geq & ~ \exp(-\| A x \|_2) \\
\geq & ~ \exp(-R^2)
\end{align*}
the 1st step follows from simple algebra, the 2nd step follows from definition of $\ell_{\infty}$ norm, the 3rd step follows from Fact~\ref{fac:vector_norm}.

\end{proof}

\subsection{Upper bound on \texorpdfstring{$M$}{}}\label{sec:newton:M}

\begin{lemma}\label{lem:M}
If the following conditions holds
\begin{itemize}
     \item $\| A \| \leq R$.
    \item $\| x \|_2 \leq R$.
    \item Let $H$ denote the hessian of loss function $L$.
    \item  $\| H(x) - H(y) \| \leq \beta^{-2} n^{1.5} \exp(20R^2) \cdot \| x- y \|_2$ (Lemma~\ref{lem:lipschitz})
\end{itemize}
Then, we have
\begin{align*}
    M \leq n^{1.5} \exp(30 R^2).
\end{align*}    
\end{lemma}
\begin{proof}
It follows from Lemma~\ref{lem:beta}.
\end{proof}

\section{Main Result}\label{sec:result}

\begin{algorithm}[!ht]\caption{Here, we present our main algorithm in an informal way.}\label{alg:main}
\begin{algorithmic}[1]
\Procedure{IterativeSoftmaxRegression}{$A \in \R^{n \times d},b \in \R^n,w \in \R^n, \epsilon, \delta$} \Comment{Theorem~\ref{thm:main_formal}} 
    \State We choose $x_0$ (suppose it satisfies Definition~\ref{def:f_ass})
    \State We use $T \gets \log( \| x_0 - x^* \|_2 / \epsilon )$ to denote the number of iterations.
    \For{$t=0 \to T$} 
        \State $D \gets B_{\diag}(x_t) + \diag(w \circ w)$ 
        \State $\wt{D} \gets \textsc{SubSample}(D,A,\epsilon_1 = \Theta(1), \delta_1 = \delta/T)$ \Comment{Lemma~\ref{lem:subsample}}
        \State $g \gets A^\top (f(x_t) \langle c(x_t) , f(x_t) \rangle + \diag(f(x_t)) c(x_t) )$
        \State $\wt{H} \gets A^\top \wt{D} A$ 
        \State $x_{t+1} \gets x_t + \wt{H}^{-1} g$ 
    \EndFor
    \State $\wt{x}\gets x_{T+1}$
    \State \Return $\wt{x}$
\EndProcedure
\end{algorithmic}
\end{algorithm}

\begin{theorem}[ 
]\label{thm:main_formal}
Suppose we have matrix $A \in \R^{n \times d}$, and vectors $b,w \in \R^n$. 
And we have the following
\begin{itemize}
    \item Define $f(x):=\langle \exp(Ax), \mathbf{1}_n \rangle^{-1} \exp(Ax)$.
    \item Define $x^*$ as the optimal solution of 
    \begin{align*}
    \min_{x \in \R^d} 0.5 \| f(Ax) - b \|_2^2 + 0.5 \| \diag(w) A x \|_2^2
    \end{align*}
    for which, 
    \begin{itemize}
        \item $g(x^*) = {\bf 0}_d$.
        \item $\| x^* \|_2 \leq R$.
    \end{itemize}
    \item Define $R \geq 10$ be a positive scalar. 
    \item It holds that $\| A \| \leq R$ 
    \item It holds that $b \geq {\bf 0}_n$, and $\| b \|_1 \leq 1$.

    \item It holds that $w_{i}^2 \geq 100 + l/\sigma_{\min}(A)^2$ for all $i \in [n]$

    \item It holds that $M = n^{1.5} \exp(30R^2)$.

    \item Let $x_0$ denote an initial point for which it holds that $M \| x_0 - x^* \|_2 \leq 0.1 l$.
\end{itemize}
Then for any accuracy parameter $\epsilon \in (0,0.1)$ and failure probability $\delta \in (0,0.1)$, there exists a randomized algorithm (Algorithm~\ref{alg:main}) such that, with probability at least $1-\delta$, it runs $T = \log(\| x_0 - x^* \|_2/ \epsilon)$ iterations and outputs a vector $\wt{x} \in \R^d$ such that
\begin{align*}
\| \wt{x} - x^* \|_2 \leq \epsilon,
\end{align*}
and the time cost per iteration is
\begin{align*}
O( (\nnz(A) + d^{\omega} ) \cdot \poly(\log(n/\delta)). 
\end{align*}

Here $\omega$ denote the exponent of matrix multiplication. Currently $\omega \approx 2.373$ \cite{w12,lg14,aw21}.  
 
\end{theorem}

\begin{proof}

    It follows from combining Lemma~\ref{lem:convex}, Lemma~\ref{lem:newton_induction}, Lemma~\ref{lem:subsample}, Lemma~\ref{lem:lipschitz} and
    Lemma~\ref{lem:one_step_shrinking}.

{\bf Proof of Upper bound on $M$.}

It follows from Lemma~\ref{lem:M}.

{\bf Proof of Hessian is PD.}

This follows from Lemma~\ref{lem:convex}.

{\bf Proof of Hessian is Lipschitz.}

This follows from Lemma~\ref{lem:lipschitz}.

{\bf Proof of Cost per iteration.}

This follows from Lemma~\ref{lem:subsample}.

{\bf Proof of Convergence per Iteration.}

By Lemma~\ref{lem:one_step_shrinking}, we have
\begin{align*}
    \|x_k - x^*\|_2 \le 0.4 \cdot \|x_{k-1} - x^*\|_2.
\end{align*}

{\bf Proof of Number of Iterations.}
    After $T$ iterations, we have
    \begin{align*}
    \| x_T - x^* \|_2 \leq 0.4^T \cdot \| x_0 - x^* \|_2
    \end{align*}
    By choice of $T$, we get the desired bound. The failure probability is following from union bound over $T$ iterations.
\end{proof}

%% file: main.bbl
\newcommand{\etalchar}[1]{$^{#1}$}
\begin{thebibliography}{KGW{\etalchar{+}}23}

\bibitem[ALS{\etalchar{+}}22]{als+22}
Josh Alman, Jiehao Liang, Zhao Song, Ruizhe Zhang, and Danyang Zhuo.
\newblock Bypass exponential time preprocessing: Fast neural network training
  via weight-data correlation preprocessing.
\newblock {\em arXiv preprint arXiv:2211.14227}, 2022.

\bibitem[Ans00]{a00}
Kurt~M Anstreicher.
\newblock The volumetric barrier for semidefinite programming.
\newblock {\em Mathematics of Operations Research}, 2000.

\bibitem[AS23]{as23}
Josh Alman and Zhao Song.
\newblock Fast attention requires bounded entries.
\newblock {\em arXiv preprint arXiv:2302.13214}, 2023.

\bibitem[AW21]{aw21}
Josh Alman and Virginia~Vassilevska Williams.
\newblock A refined laser method and faster matrix multiplication.
\newblock In {\em Proceedings of the 2021 ACM-SIAM Symposium on Discrete
  Algorithms (SODA)}, pages 522--539. SIAM, 2021.

\bibitem[AZLS19a]{als19_dnn}
Zeyuan Allen-Zhu, Yuanzhi Li, and Zhao Song.
\newblock A convergence theory for deep learning via over-parameterization.
\newblock In {\em International Conference on Machine Learning}, pages
  242--252. PMLR, 2019.

\bibitem[AZLS19b]{als19_rnn}
Zeyuan Allen-Zhu, Yuanzhi Li, and Zhao Song.
\newblock On the convergence rate of training recurrent neural networks.
\newblock {\em Advances in neural information processing systems}, 32, 2019.

\bibitem[BCE{\etalchar{+}}23]{bce+23}
S{\'e}bastien Bubeck, Varun Chandrasekaran, Ronen Eldan, Johannes Gehrke, Eric
  Horvitz, Ece Kamar, Peter Lee, Yin~Tat Lee, Yuanzhi Li, Scott Lundberg,
  et~al.
\newblock Sparks of artificial general intelligence: Early experiments with
  gpt-4.
\newblock {\em arXiv preprint arXiv:2303.12712}, 2023.

\bibitem[BMR{\etalchar{+}}20]{bmr+20}
Tom Brown, Benjamin Mann, Nick Ryder, Melanie Subbiah, Jared~D Kaplan, Prafulla
  Dhariwal, Arvind Neelakantan, Pranav Shyam, Girish Sastry, Amanda Askell,
  et~al.
\newblock Language models are few-shot learners.
\newblock {\em Advances in neural information processing systems},
  33:1877--1901, 2020.

\bibitem[BPSW21]{bpsw21}
Jan van~den Brand, Binghui Peng, Zhao Song, and Omri Weinstein.
\newblock Training (overparametrized) neural networks in near-linear time.
\newblock In {\em ITCS}, 2021.

\bibitem[Bra20]{b20}
Jan van~den Brand.
\newblock A deterministic linear program solver in current matrix
  multiplication time.
\newblock In {\em Proceedings of the Fourteenth Annual ACM-SIAM Symposium on
  Discrete Algorithms (SODA)}, pages 259--278. SIAM, 2020.

\bibitem[BSZ23]{bsz23}
Jan van~den Brand, Zhao Song, and Tianyi Zhou.
\newblock Algorithm and hardness for dynamic attention maintenance in large
  language models.
\newblock {\em arXiv preprint arXiv:2304.02207}, 2023.

\bibitem[CCLY19]{ccly19}
Michael~B Cohen, Ben Cousins, Yin~Tat Lee, and Xin Yang.
\newblock A near-optimal algorithm for approximating the john ellipsoid.
\newblock In {\em Conference on Learning Theory}, pages 849--873. PMLR, 2019.

\bibitem[Cha22]{cha22}
ChatGPT.
\newblock Optimizing language models for dialogue.
\newblock {\em OpenAI Blog}, November 2022.

\bibitem[CLP{\etalchar{+}}21]{clp+21}
Beidi Chen, Zichang Liu, Binghui Peng, Zhaozhuo Xu, Jonathan~Lingjie Li, Tri
  Dao, Zhao Song, Anshumali Shrivastava, and Christopher Re.
\newblock Mongoose: A learnable lsh framework for efficient neural network
  training.
\newblock In {\em International Conference on Learning Representations}, 2021.

\bibitem[CLS19]{cls19}
Michael~B Cohen, Yin~Tat Lee, and Zhao Song.
\newblock Solving linear programs in the current matrix multiplication time.
\newblock In {\em STOC}, 2019.

\bibitem[CND{\etalchar{+}}22]{cnd+22}
Aakanksha Chowdhery, Sharan Narang, Jacob Devlin, Maarten Bosma, Gaurav Mishra,
  Adam Roberts, Paul Barham, Hyung~Won Chung, Charles Sutton, Sebastian
  Gehrmann, et~al.
\newblock Palm: Scaling language modeling with pathways.
\newblock {\em arXiv preprint arXiv:2204.02311}, 2022.

\bibitem[DCLT18]{dclt18}
Jacob Devlin, Ming-Wei Chang, Kenton Lee, and Kristina Toutanova.
\newblock Bert: Pre-training of deep bidirectional transformers for language
  understanding.
\newblock {\em arXiv preprint arXiv:1810.04805}, 2018.

\bibitem[DLS23]{dls23}
Yichuan Deng, Zhihang Li, and Zhao Song.
\newblock An improved sample complexity for rank-1 matrix sensing.
\newblock {\em arXiv preprint arXiv:2303.06895}, 2023.

\bibitem[DLY21]{dly21}
Sally Dong, Yin~Tat Lee, and Guanghao Ye.
\newblock A nearly-linear time algorithm for linear programs with small
  treewidth: a multiscale representation of robust central path.
\newblock In {\em Proceedings of the 53rd Annual ACM SIGACT Symposium on Theory
  of Computing}, pages 1784--1797, 2021.

\bibitem[DMS23]{dms23}
Yichuan Deng, Sridhar Mahadevan, and Zhao Song.
\newblock Randomized and deterministic attention sparsification algorithms for
  over-parameterized feature dimension.
\newblock {\em arxiv preprint: arxiv 2304.03426}, 2023.

\bibitem[DSW22]{dsw22}
Yichuan Deng, Zhao Song, and Omri Weinstein.
\newblock Discrepancy minimization in input-sparsity time.
\newblock {\em arXiv preprint arXiv:2210.12468}, 2022.

\bibitem[GMS23]{gms23}
Yeqi Gao, Sridhar Mahadevan, and Zhao Song.
\newblock An over-parameterized exponential regression.
\newblock {\em arXiv preprint arXiv:2303.16504}, 2023.

\bibitem[GS22]{gs22}
Yuzhou Gu and Zhao Song.
\newblock A faster small treewidth sdp solver.
\newblock {\em arXiv preprint arXiv:2211.06033}, 2022.

\bibitem[GSYZ23]{gsyz23}
Yuzhou Gu, Zhao Song, Junze Yin, and Lichen Zhang.
\newblock Low rank matrix completion via robust alternating minimization in
  nearly linear time.
\newblock {\em arXiv preprint arXiv:2302.11068}, 2023.

\bibitem[HJS{\etalchar{+}}22]{hjs+22}
Baihe Huang, Shunhua Jiang, Zhao Song, Runzhou Tao, and Ruizhe Zhang.
\newblock Solving sdp faster: A robust ipm framework and efficient
  implementation.
\newblock In {\em 2022 IEEE 63rd Annual Symposium on Foundations of Computer
  Science (FOCS)}, pages 233--244. IEEE, 2022.

\bibitem[HWL21]{hwl21}
Weihua He, Yongyun Wu, and Xiaohua Li.
\newblock Attention mechanism for neural machine translation: A survey.
\newblock In {\em 2021 IEEE 5th Information Technology, Networking, Electronic
  and Automation Control Conference (ITNEC)}, volume~5, pages 1485--1489. IEEE,
  2021.

\bibitem[JKL{\etalchar{+}}20]{jkl+20}
Haotian Jiang, Tarun Kathuria, Yin~Tat Lee, Swati Padmanabhan, and Zhao Song.
\newblock A faster interior point method for semidefinite programming.
\newblock In {\em 2020 IEEE 61st annual symposium on foundations of computer
  science (FOCS)}, pages 910--918. IEEE, 2020.

\bibitem[JLSW20]{jlsw20}
Haotian Jiang, Yin~Tat Lee, Zhao Song, and Sam Chiu-wai Wong.
\newblock An improved cutting plane method for convex optimization,
  convex-concave games and its applications.
\newblock In {\em STOC}, 2020.

\bibitem[JLSZ23]{jlsz23}
Haotian Jiang, Yin~Tat Lee, Zhao Song, and Lichen Zhang.
\newblock Convex minimization with integer minima in $\widetilde{O}(n^4)$ time.
\newblock {\em arXiv preprint arXiv:2304.03426}, 2023.

\bibitem[JSWZ21]{jswz21}
Shunhua Jiang, Zhao Song, Omri Weinstein, and Hengjie Zhang.
\newblock Faster dynamic matrix inverse for faster lps.
\newblock In {\em STOC}, 2021.

\bibitem[KGW{\etalchar{+}}23]{kgw+23}
John Kirchenbauer, Jonas Geiping, Yuxin Wen, Jonathan Katz, Ian Miers, and Tom
  Goldstein.
\newblock A watermark for large language models.
\newblock {\em arXiv preprint arXiv:2301.10226}, 2023.

\bibitem[KKL20]{kkl20}
Nikita Kitaev, {\L}ukasz Kaiser, and Anselm Levskaya.
\newblock Reformer: The efficient transformer.
\newblock {\em arXiv preprint arXiv:2001.04451}, 2020.

\bibitem[LG14]{lg14}
Fran{\c{c}}ois Le~Gall.
\newblock Powers of tensors and fast matrix multiplication.
\newblock In {\em Proceedings of the 39th international symposium on symbolic
  and algebraic computation}, pages 296--303, 2014.

\bibitem[LLR23]{llr23}
Yuchen Li, Yuanzhi Li, and Andrej Risteski.
\newblock How do transformers learn topic structure: Towards a mechanistic
  understanding.
\newblock {\em arXiv preprint arXiv:2303.04245}, 2023.

\bibitem[LSZ19]{lsz19}
Yin~Tat Lee, Zhao Song, and Qiuyi Zhang.
\newblock Solving empirical risk minimization in the current matrix
  multiplication time.
\newblock In {\em Conference on Learning Theory (COLT)}, pages 2140--2157.
  PMLR, 2019.

\bibitem[LSZ23]{lsz23}
Zhihang Li, Zhao Song, and Tianyi Zhou.
\newblock Solving regularized exp, cosh and sinh regression problems.
\newblock {\em arXiv preprint, 2303.15725}, 2023.

\bibitem[MMS{\etalchar{+}}19]{mms+19}
Louis Martin, Benjamin Muller, Pedro Javier~Ortiz Suarez, Yoann Dupont, Laurent
  Romary, Eric~Villemonte de~La~Clergerie, Djame Seddah, and Benoit Sagot.
\newblock Camembert: a tasty french language model.
\newblock {\em arXiv preprint arXiv:1911.03894}, 2019.

\bibitem[Ope23]{o23}
OpenAI.
\newblock Gpt-4 technical report.
\newblock {\em arXiv preprint arXiv:2303.08774}, 2023.

\bibitem[QSZ23]{qsz23}
Lianke Qin, Zhao Song, and Ruizhe Zhang.
\newblock A general algorithm for solving rank-one matrix sensing.
\newblock {\em arXiv preprint arXiv:2303.12298}, 2023.

\bibitem[QSZZ23]{qszz23}
Lianke Qin, Zhao Song, Lichen Zhang, and Danyang Zhuo.
\newblock An online and unified algorithm for projection matrix vector
  multiplication with application to empirical risk minimization.
\newblock In {\em AISTATS}, 2023.

\bibitem[RNS{\etalchar{+}}18]{rns+18}
Alec Radford, Karthik Narasimhan, Tim Salimans, Ilya Sutskever, et~al.
\newblock Improving language understanding by generative pre-training.
\newblock {\em .}, 2018.

\bibitem[RWC{\etalchar{+}}19]{rwc+19}
Alec Radford, Jeffrey Wu, Rewon Child, David Luan, Dario Amodei, Ilya
  Sutskever, et~al.
\newblock Language models are unsupervised multitask learners.
\newblock {\em OpenAI blog}, 1(8):9, 2019.

\bibitem[SY21]{sy21}
Zhao Song and Zheng Yu.
\newblock Oblivious sketching-based central path method for linear programming.
\newblock In {\em International Conference on Machine Learning}, pages
  9835--9847. PMLR, 2021.

\bibitem[SYYZ22]{syyz22}
Zhao Song, Xin Yang, Yuanyuan Yang, and Tianyi Zhou.
\newblock Faster algorithm for structured john ellipsoid computation.
\newblock {\em arXiv preprint arXiv:2211.14407}, 2022.

\bibitem[SZKS21]{szks21}
Charlie Snell, Ruiqi Zhong, Dan Klein, and Jacob Steinhardt.
\newblock Approximating how single head attention learns.
\newblock {\em arXiv preprint arXiv:2103.07601}, 2021.

\bibitem[SZZ21]{szz21}
Zhao Song, Lichen Zhang, and Ruizhe Zhang.
\newblock Training multi-layer over-parametrized neural network in subquadratic
  time.
\newblock {\em arXiv preprint arXiv:2112.07628}, 2021.

\bibitem[UAS{\etalchar{+}}20]{uas20}
Mohd Usama, Belal Ahmad, Enmin Song, M~Shamim Hossain, Mubarak Alrashoud, and
  Ghulam Muhammad.
\newblock Attention-based sentiment analysis using convolutional and recurrent
  neural network.
\newblock {\em Future Generation Computer Systems}, 113:571--578, 2020.

\bibitem[VKB23]{vkb23}
Nikhil Vyas, Sham Kakade, and Boaz Barak.
\newblock Provable copyright protection for generative models.
\newblock {\em arXiv preprint arXiv:2302.10870}, 2023.

\bibitem[VSP{\etalchar{+}}17]{vsp+17}
Ashish Vaswani, Noam Shazeer, Niki Parmar, Jakob Uszkoreit, Llion Jones,
  Aidan~N Gomez, {\L}ukasz Kaiser, and Illia Polosukhin.
\newblock Attention is all you need.
\newblock {\em Advances in neural information processing systems}, 30, 2017.

\bibitem[Wil12]{w12}
Virginia~Vassilevska Williams.
\newblock Multiplying matrices faster than coppersmith-winograd.
\newblock In {\em Proceedings of the forty-fourth annual ACM symposium on
  Theory of computing}, pages 887--898, 2012.

\bibitem[Zha22]{z22}
Lichen Zhang.
\newblock Speeding up optimizations via data structures: Faster search, sample
  and maintenance.
\newblock Master's thesis, Carnegie Mellon University, 2022.

\bibitem[ZHDK23]{zhdk23}
Amir Zandieh, Insu Han, Majid Daliri, and Amin Karbasi.
\newblock Kdeformer: Accelerating transformers via kernel density estimation.
\newblock {\em arXiv preprint arXiv:2302.02451}, 2023.

\bibitem[ZKV{\etalchar{+}}20]{zkv+20}
Jingzhao Zhang, Sai~Praneeth Karimireddy, Andreas Veit, Seungyeon Kim, Sashank
  Reddi, Sanjiv Kumar, and Suvrit Sra.
\newblock Why are adaptive methods good for attention models?
\newblock {\em Advances in Neural Information Processing Systems},
  33:15383--15393, 2020.

\bibitem[ZRG{\etalchar{+}}22]{zrg+22}
Susan Zhang, Stephen Roller, Naman Goyal, Mikel Artetxe, Moya Chen, Shuohui
  Chen, Christopher Dewan, Mona Diab, Xian Li, Xi~Victoria Lin, et~al.
\newblock Opt: Open pre-trained transformer language models.
\newblock {\em arXiv preprint arXiv:2205.01068}, 2022.

\end{thebibliography}
